\documentclass[journal]{IEEEtran}

\usepackage{times}
\usepackage{epsfig}
\usepackage{graphicx,times}
\usepackage{subfigure}  
\usepackage{amsmath}
\usepackage{amssymb}
\usepackage{booktabs}
\usepackage{epstopdf}
\usepackage[dvipsnames, svgnames, x11names]{xcolor}
\usepackage{multirow}

\newcommand{\squishlist}{
	\begin{list}{$\bullet$}
		{ \setlength{\itemsep}{0pt}
			\setlength{\parsep}{3pt}
			\setlength{\topsep}{3pt}
			\setlength{\partopsep}{0pt}
			\setlength{\leftmargin}{1.5em}
			\setlength{\labelwidth}{1em}
			\setlength{\labelsep}{0.5em}}
	}
	
	\newcommand{\squishend}{
\end{list}  }

\ifCLASSINFOpdf
\else
\fi
\begin{document}
%
\title{DR-GAN: Distribution Regularization for Text-to-Image  Generation}
%
%
%

	\author{Hongchen Tan,   Xiuping Liu,   Baocai Yin  and Xin Li*,~\IEEEmembership{Senior Member, IEEE}

	\thanks{(Corresponding author: Xin Li)
		
			Hongchen Tan and  Baocai Yin are  with Artificial Intelligence Research Institute, Beijing University of Technology, Beijing 100124, China (e-mail: tanhongchenphd@bjut.edu.cn; ybc@bjut.edu.cn).

            Xin Li is with School of Electrical Engineering \& Computer Science, and Center for Computation \& Technology, Louisiana State University, Baton Rouge (LA) 70808, United States of America (e-mail: xin.shane.li@ieee.org).

            Xiuping Liu is with School of Mathematical Sciences, Dalian University of Technology, Dalian 116024, China (xpliu@dlut.edu.cn.).

	}
	\thanks{}
	\thanks{}}

\markboth{Journal of \LaTeX\ Class Files,~Vol.~14, No.~8, August~2015}%
{Shell \MakeLowercase{\textit{et al.}}: Bare Demo of IEEEtran.cls for IEEE Journals}
%

\maketitle

\begin{abstract}
This paper presents a new Text-to-Image  generation model, named Distribution Regularization Generative Adversarial Network (DR-GAN), to generate images from text descriptions from improved distribution learning. 
In DR-GAN, we  introduce two  novel  modules: a Semantic  Disentangling Module (SDM) and a Distribution Normalization Module (DNM). 
SDM combines the spatial self-attention mechanism and a new Semantic Disentangling Loss (SDL) to help the generator distill key semantic information for  the image generation.
DNM uses a Variational Auto-Encoder (VAE) to normalize and denoise the image latent distribution, which can help the    discriminator better distinguish synthesized images from real images. 
DNM also adopts a Distribution Adversarial Loss (DAL)  to guide the generator to  align with normalized real image distributions in the latent space.
Extensive experiments on two public datasets demonstrated that our DR-GAN achieved a competitive  performance  in the Text-to-Image  task.	
The code link: https://github.com/Tan-H-C/DR-GAN-Distribution-Regularization-for-Text-to-Image-Generation
\end{abstract}

\section{Introduction}~\label{Introduction}

Generating photographic images from text descriptions (known as Text-to-Image Generation, T2I) is a challenging cross-modal generation technique that is a core component in many computer vision tasks 
such as Image Editing~\cite{LisaiACMMM2020, yahuiACMMM2020}, Story Visualization~\cite{StoryGAN2019}, and Multimedia Retrieval~\cite{Jiuxiang2018}. 
Compared with the image generation~\cite{2020MixNMatch, Isola2016Image, Jun-Yan2017iccv} and image processing~\cite{2017Compressed, DBLPDEEP, Kalayeh2018Human} tasks between the same mode,  it is difficult   to  build  the  heterogeneous  semantic bridge between text and image~\cite{Mingkuan_2018, wang2020context, 9210842}.  
Many state-of-the-art T2I algorithms~\cite{Minfeng2019, Wenb19cvpr, 2019Tingting, Guojun2019, Bowen2020, Tingtingnips2019} first extract text features, then use Generative Adversarial Networks (GANs)~\cite{Goodfellow2014Generative} to generate the corresponding image. 
Their essence is to map a text feature distribution to the image distribution. 
However, two factors prevent GAN-based T2I  methods  from capturing real image distribution: 
(1) The abstract  and ambiguity of text descriptions  make the  generator difficult  capture  the key semantic information for  image  generation~\cite{Mingkuan_2018, Yuxin_2017};
(2) The diversity of visual information  makes  the distribution of images complex so that it  is difficult  for the   GAN-based T2I models  to capture  the real image distribution from  text  feature  distribution~\cite{Generative_2020}.
Thus, this work explores better distribution learning strategies to enhance GAN-based T2I models.

In multi-modal perceptual information, the semantics  of  the text description is usually abstract and ambiguous; Image information is usually concrete and has a lot of  spatial structure information. Text and Image information are expressed in different patterns, which makes it difficult to achieve semantic correlation based on feature vectors or tensors. Thus, it is difficult for the generator to accurately capture key semantics  from text descriptions  for  image generation.  Since this, in the intermediate stage of generation, the   image  features  contain  a lot of non-key semantics.  Such inaccurate semantics often leads to ineffective image distribution generation, and then the generated images are often semantically inconsistent, chaos structure and details and so on. 
To alleviate this issue, our first strategy is to design an information disentangling mechanism on the intermediate  feature,  to better distill key information before performing cross-modal distribution learning. 


In addition,  images  often contain  diverse visual information, messy background,  and   other  non-key  visual  information. 
Their image latent distribution is often complex. And, the distribution of images is difficult to model explicitly~\cite{Generative_2020}. 
This means that we cannot directly and explicitly learn the target image distribution from the text features.  
As an outstanding image generation model, GANs~\cite{Goodfellow2014Generative}  learn the target data distribution  implicitly by sampling data from the True or Fake data  distribution.
However, such complex  image   distribution  makes it difficult for the discriminator in GANs to distinguish whether the current input image is sampled from the real image distribution or generated image distribution.
So, our second strategy is to design an effective distribution normalization mechanism to normalize   the image latent  distribution. The  mechanism  aims to  help the discriminator better learn the distribution decision boundary between the generated versus real image.

Based on the above two strategies, we built a new Text-to-Image generation model, Distribution Regularization Generative Adversarial Networks (DR-GAN). 
DR-GAN contains two novel modules: Semantic  Disentangling Module (SDM) and Distribution Normalization  Module (DNM).
In SDM, we introduce a spatial self-attention mechanism and  propose  a new  Semantic  Disentangling Loss (SDL) to help the generator  better distill key information from texts and images in capturing the image  distribution  process. 
In DNM, we introduce a Variational Auto-Encoder (VAE)~\cite{kingma2014autoencoding}  into   GAN-based T2I methods to normalize image distributions in the latent space. 
We also propose a Distribution Adversarial Loss (DAL) to align the learned distribution with the  real distribution  in the normalized  latent space.
With DNM and SDM, our DR-GAN can generate a  image latent  distribution that better matches with the real image distribution, and generate higher-quality images.   
The \textbf{main contributions} are summarized as follows:

\squishlist 
\item[(i)] We propose a  Semantic  Disentangling Module (SDM) to help the generator distill key information (and filter out the non-key information) from both  text and image features.

\item[(ii)] We design a new Distribution Normalization Module (DNM) which introduces the VAE into the GAN-based T2I pipeline so that it can more effectively normalize  and  denoise image  latent distributions.  

\item[(iii)] Extensive experimental results and analysis show the efficacy of DR-GAN on two benchmarks: CUB-Bird \cite{WahCUB_200_2011} and large-scale MS-COCO \cite{Lin2014Microsoft} over four metrics.
\squishend

\section{Related  Work}

\subsection{GANs in Text-to-Image Generation}

With the recent successes of  GANs~\cite{Goodfellow2014Generative}, a large number of GAN-based  T2I  methods~\cite{Reed2016Generative, Han2017StackGAN, Han2018StackGANn,  Xu2017AttnGAN, Wenb19cvpr, 2019Tingting, Bowen2020, thc2019, Guojun2019, Minfeng2019, Tingtingnips2019} have boosted the performance  of the T2I task. 
Reed et al.~\cite{Reed2016Generative} first introduced the adversarial process to generate images from text descriptions.  
However, they can only generate images with $64 \times 64$ resolution. And, the quality of the images is not good.
Then, StackGAN/StackGAN++~\cite{Han2017StackGAN, Han2018StackGANn} and HDGAN~\cite{Zhang2018Photographic}   adopted multi-stage generation patterns to progressively enhance the quality and detail of synthesized images. 
Thereafter, the multi-stage generation  framework has been widely used in GAN-based T2I  methods. 
Based on the multi-stage generation  framework,  AttnGAN~\cite{Xu2017AttnGAN}, DMGAN~\cite{Minfeng2019},  and CPGAN~\cite{Jiadongeccv2020} adopted the word-level or the object-level attention mechanisms to help the generator enhance  local regions'  or  objects' semantics. 
MirrorGAN~\cite{2019Tingting} combined   the  Text-to-Image generation  and Image-to-Text   generation  to improve  the global semantic consistency between text description and  the generated image. 
SDGAN~\cite{Guojun2019} and SEGAN~\cite{thc2019} combined  the Siamese Network and the contrastive  loss  to   enhance the semantics of the synthesized  image. 
Like these T2I methods, we also adopt the multi-stage generation framework to build  our DR-GAN.  
But different from them, we help the  generator to better distill key information for  distribution learning.

\subsection{GANs in distribution learning} 
The GANs~\cite{Goodfellow2014Generative},  as a latent distribution learning strategy, has been widely adopted in various generation tasks~\cite{Reed2016Generative, Assafcvpr2020, Jun-Yan2017iccv, Phillip2017cvpr, yahuiACMMM2020, StoryGAN2019}. 
However, GANs tend to suffer from unstable training, mode collapse, and uncontrollable problem, etc., and they hinder   GANs from effectively modeling the real data distribution. 
Recently, many  approaches~\cite{Nowozinnips2016, Han2018StackGANn, Cheiclr2017, maoiccv2017, Miyato2018,  Han2017StackGAN} have been exploited to overcome these issues. 
LSGANs~\cite{maoiccv2017} overcome the vanishing gradient problem by introducing a least square loss to replace the entropy loss.   
WGAN~\cite{MartinICML2017} introduced the Earth Mover (EM) distance to improve the stability of learning distribution and provide meaningful learning curves useful for hyperparameter search and debugging. 
MDGAN~\cite{Cheiclr2017} improved the stability of distribution learning by utilizing an encoder $E(x): x \rightarrow z$ to produce the latent variable $z$ for the generator $G$. 
SN-GANs~\cite{Miyato2018} proposed a novel weight normalization method named spectral normalization to better stabilize the training of the discriminator.  
F-GAN~\cite{Nowozinnips2016} adopted  the Kullback-Leibler (KL) divergence to help  align the generated image distribution with the real data distribution. 
In the T2I task, many GAN-based methods introduce various strategies such as multi-stage generation pattern~\cite{Han2018StackGANn, Han2017StackGAN}, attention mechanism~\cite{Xu2017AttnGAN, Minfeng2019}, and cycle consistent mechanism~\cite{2019Tingting} to help match
synthesized distribution with the real  image distribution.
However,  diverse visual information, messy background,  and   other  non-key  visual  information  in  images usually make the image distribution  complicated.  It makes distribution learning  more  difficult.  
Thus, our idea is to explore an effective distribution normalization strategy   to overcome the challenge.

\section{DR-GAN for text-to-image generation }

\begin{figure}[h!tb]
	\centering
	\includegraphics[scale=0.43]{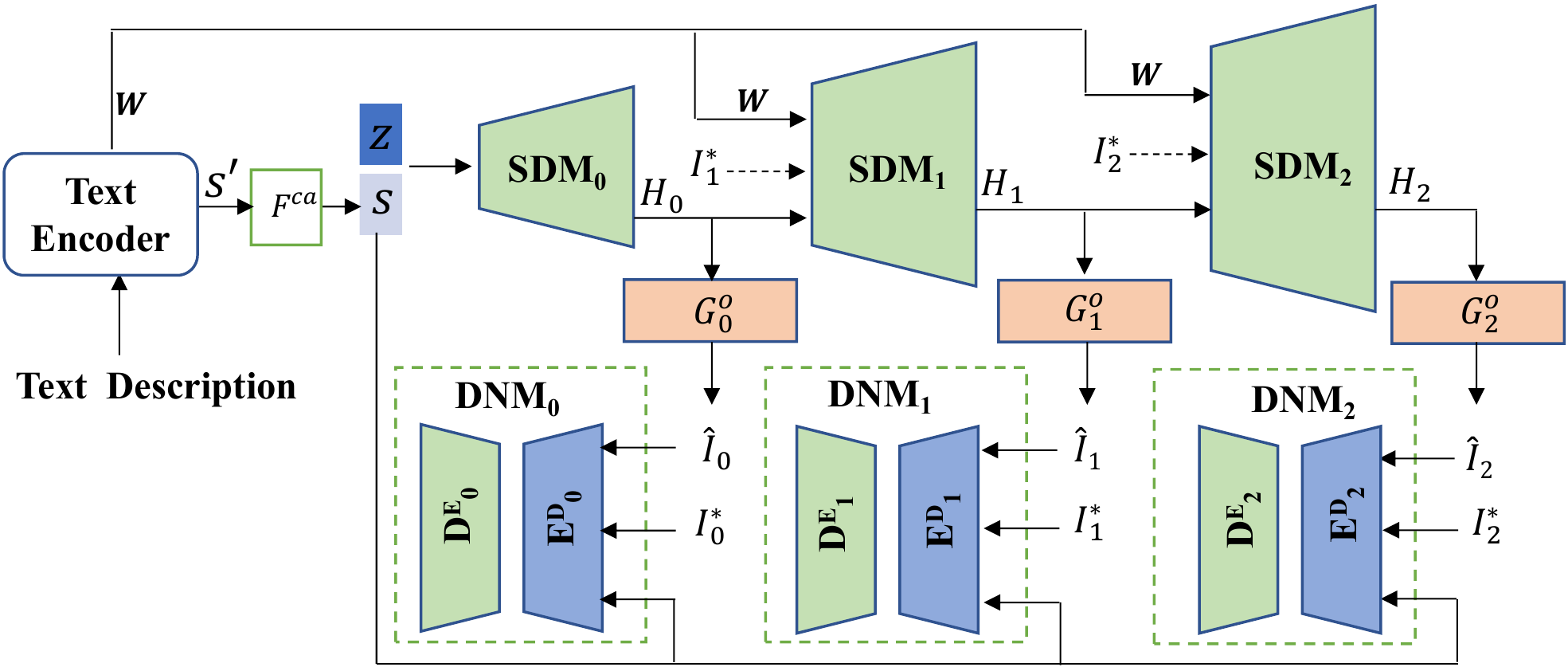}
	\caption{The framework of   proposed Distribution  Regularization  Generative Adversarial Network (DR-GAN).} 
	\label{TNNLS-1} 
\end{figure}

\subsection{Overview} Most recent GAN-based T2I methods~\cite{Xu2017AttnGAN, Tobias2019, Wenb19cvpr, 2019Tingting, thc2019, Minfeng2019, Guojun2019} adopted a multi-stage generation framework to progressively map the text  embedding distribution to the image distribution, to synthesize high-quality  images. 
Like all these methods, we also adopt such a generation pattern from AttnGAN~\cite{Xu2017AttnGAN} as our baseline to build the DR-GAN. 

As shown in Fig.~\ref{TNNLS-1}, DR-GAN has a Text Encoder~\cite{Xu2017AttnGAN}, a conditioning augmentation module~\cite{Han2017StackGAN} $F^{ca}$, $m$ generation modules $G_{i}^{o}$, ($\emph{i} =0, 1, 2, \ldots, m-1$), and 
two new designs: 
$m$ Semantic Disentangling Modules (SDMs) $SDM_{i}$, and $m$ Distribution Normalization Modules (DNMs) $DNM_{i}$, $\emph{i} =0, 1, 2, \ldots, m-1$.  

The Text Encoder transforms the input text description (a single sentence) into the sentence feature $s'$ and word features $W$. 
The $F^{ca}$~\cite{Han2017StackGAN} converts a sentence feature $s'$ to a conditioning sentence feature $s$. 
The SDM   distills key information from the text or image features, in  the intermediate stage of generation, for  better  approximating the real image  distribution. 
In the \emph{testing stage}, 
these SDMs take  noise $z \sim N(0, 1)$,  the sentence  feature  $s$,  and  the word  features $W$ to produce a series of hidden features $H_{i}$ ($\emph{i} =0, 1,2,\cdots,m-1$); 
while in the \emph{training stage}, 
besides $z$,  $s$ and $W$, the $i$-th $SDM_{i}$ also takes the $i$-th scale  real image, $I^{*}_i$, as input to generate $H_i$.
Then, $G_{i}^{o}$ takes $H_{i}$ to generate 
$i$-th scaled images $\hat{I}_{i}$. 
The $DNM_{i}$ contains a Variational Auto-Encoder (VAE) module and a discriminator: the former normalizes image latent distributions, and the latter distinguishes between  the  real image and the synthesized image.  
The generation stage information flow is formulated as

\begin{small}
	\begin{equation}\label{overview} 
		\begin{aligned}
			&SDM_0: H_0=SDM_0(z,F^{ca}(s'), I_{0}^{*})\,\,\,  \textrm{in Training Stage};\\
			&SDM_i: H_i=SDM_i(H_{i-1}, W,  I_{i}^{*})\,\,\,  \textrm{in Training Stage};\\
			&SDM_0: H_0=SDM_0(z,F^{ca}(s'))\,\,\,  \textrm{in Testing Stage};\\
			&SDM_i: H_i=SDM_i(H_{i-1}, W)\,\,\, \textrm{in Testing Stage};\\
			&G_i^{o}: \hat{I_i}=G_i^{o}(H_i), \,\,\,  \emph{i} =0, 1, 2, \cdots, m-1.
		\end{aligned}
	\end{equation}
\end{small}
Note that the $SDM_0$    only  contains  a  series  of  convolution layers and upsampling modules.   The  proposed   Semantic  Disentangling Module (SDM)  is  adopted   in  the   $SDM_i$ $i =0, 1, 2, \cdots, m-1$.

\subsection{Semantic Disentangling Module (SDM)}~\label{DDM_Section}

\begin{figure}[h!tb]
	\centering
	\includegraphics[scale=0.22]{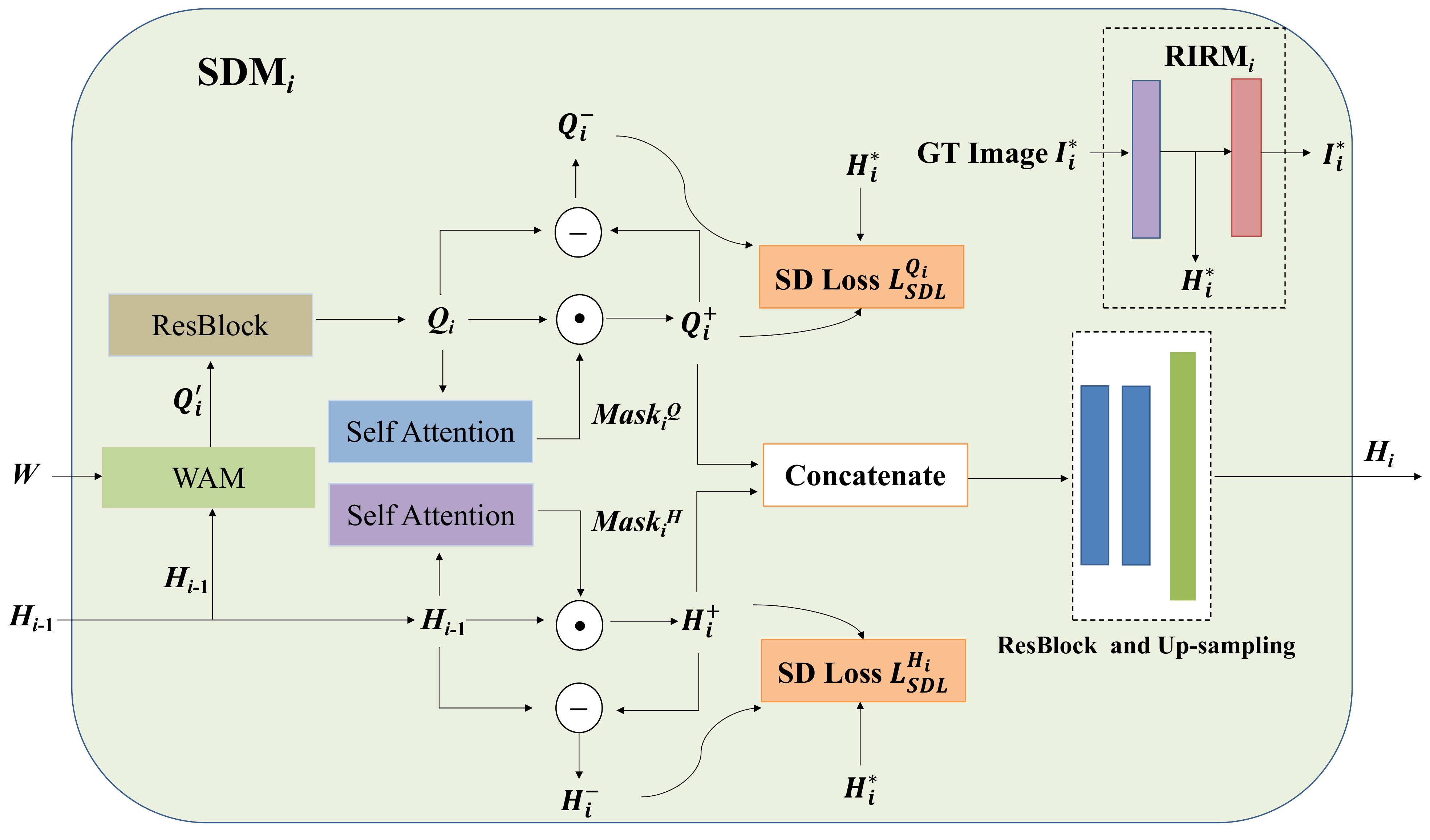}
	\caption{The architecture of the proposed Semantic  Disentangling Module (SDM). WAM: Word-level  Attention Mechanism (WAM)~\cite{Xu2017AttnGAN};  RIRM: Real Image Reconstruction Module; SD Loss: Semantic  Disentangling  Loss. } 
	\label{TNNLS-2} 
\end{figure}

Semantic  abstract  and   ambiguity of  text   description   often makes the generator difficult capture accurate semantics of the given sentence.  Such inaccurate semantics  are not conducive to learning distribution, and then lead to incorrect spatial structures and semantics in synthesized images. 
To this end, we propose a Semantic Disentangling Module (SDM) (Fig.~\ref{TNNLS-2}) to help generators suppress irrelevant spatial information and highlight relevant spatial information for generating  high-quality images.

We build our SDM based on the widely adopted  Cascaded  Attentional Generative Model (CAGM)  introduced by AttnGAN~\cite{Xu2017AttnGAN}, because the  Word-level Attention Mechanism (WAM) in CAGM can effectively enhance semantic details of the generated images.

Initially, the SDM was designed to directly extract the key  and non-key information from text features. 
Because text and image belong to the heterogeneous domain, it is difficult to achieve reasonable semantics matching in  the  feature  space~\cite{Mingkuan_2018}. 
Compared  with  the  text features,  the word-level context feature from WAM  contain  more structural  and semantic information, and semantically better match the image features.  
Besides,   the semantics   of  word-level context features  and   image features,   are   from  the input  word features $W$ and sentences  features $s$.  
Therefore,  SDM  is  designed  to  distill key information and  filter  out  non-key information, on the word-level context features  and   image features of the CAGM.

In this section, we firstly   revisit the WAM ~\cite{Xu2017AttnGAN}  to acquire the word-level context features  and image  features. 
Secondly,  we  introduce  the  Spatial Self-Attention Mechanism (SSAM) to represent  the key information and  the non-key information. 
Finally, we introduce the Semantic Disentangling Loss (SDL) to  drive  the SDM  conduct  the  semantic disentangling.

\subsubsection{Word-level Attention Mechanism (WAM)~\cite{Xu2017AttnGAN}}~\label{DDM_Section}
In cases where there is no ambiguity, we omit subscripts in this  WAM's description.
As  shown  in Fig.~\ref{TNNLS-2}, the ``WAM''  has two inputs: the word features $W\in \mathbb{R}^{D \times T}$  and the image features $H \in \mathbb{R}^{\hat{D} \times N}$ from the previous hidden layer. 

Firstly, the word features are mapped to  the  same latent semantic space as the image features,  i.e. $W'=UW$, $W'=\{w'_{i} \in \mathbb{R}^{\hat{D}} | i=1,2,\cdots,T\}$, where $U \in \mathbb{R}^{\hat{D} \times D}$ is a perceptual layer.
Each column of $H=\{h_{j} \in \mathbb{R}^{\hat{D}} | j=1,2,\cdots,N\}$ (hidden features) is a feature vector of an image's sub-region.

Secondly, for the $j^{th}$ image's sub-region, its \emph{dynamic representation} of word vectors w.r.t. $h_j$ is 
\begin{equation}\label{eq7}
	q_j=\sum_{i=1}^{T}\theta_{j,i}w_i', \quad {\rm where}\  \theta_{j,i}=\frac{exp(S_{j,i}')}{\sum_{k=1}^{T}exp(S_{j,k}')}.
\end{equation}
Here $S'_{j,i}=h^{T}_{j}w'_{i}$, and $\theta_{j,i}$ indicates the weight the model assigned to the ${i}^{th}$ word when generating the $j^{th}$ sub-region of the image.

Thirdly,  the \emph{word-level context feature} for  $H$  is denoted by   $Q'=(q_{1}, q_{2}, ..., q_{N}) \in \mathbb{R}^{\hat{D} \times N}$.

As shown in Fig.~\ref{TNNLS-2},  the WAM generates a \emph{word-level context feature} $Q'$ from the given word feature $W$ and image feature $H$. 
Here, $Q'$ is a weighted combination of word features that expresses the image feature $H$. The $Q'$ can effectively enrich the semantics of image details~\cite{Xu2017AttnGAN}. 
Due to the abstract  and   ambiguity of  text   description, the  generator is  prone to parsing wrong or inaccurate semantics. 
However,  such inaccurate semantic extraction  leads to incorrect semantics and structures in  $Q'$ and $H$. 
Thus,  it is necessary  to distill  the key information from \emph{word-level context feature}  $Q'$   and  the  intermediate  image feature $H$, for  image  generation. 
Next,  we  use Spatial Self-Attention Mechanism to represent the key and non-key features of  the   $Q'$   and  the   $H$  respectively.

\subsubsection{Spatial Self-Attention Mechanism}~\label{DDM_Section}
As shown in Fig.~\ref{TNNLS-2},  we use a Spatial Self-Attention Mechanism to represent the  key and non-key information of  the \emph{word-level context feature}  $Q'_i$   and  the  intermediate  image feature $H_{i-1}$  respectively.

Firstly, we  represent the  key and non-key information of  the  image feature $H_{i-1}$. 
The spatial attention mask $Mask_i^H$  of the feature $H_{i-1}$  is  defined as

\begin{small}
	\begin{equation}\label{Spatial_Self_Attention_Mask}
		Mask_i^H= Sig.(Conv_{1 \times 1}^2 (ReLU(Conv_{3 \times 3}^1(H_{i-1})))),
	\end{equation}
\end{small} 
where $H_{i-1}$ is followed by a $3 \times 3$ convolutional layer $Conv_{3 \times 3}^1(\cdot)$ and  a $1 \times 1$ convolutional layer $Conv_{1 \times 1}^2(\cdot)$,
and ReLU and Sigmoid (Sig.) are used as activation functions. 
We use $H^+_{i} = Mask_i^H \odot H_{i-1}$ to express the key spatial information of $H_{i-1}$, and use  $H^-_{i}=H_{i-1}-H^+_{i}$  to express the non-key information.

Secondly, we  represent the key and non-key information of the \emph{word-level context feature} $Q'_i$. Although $Q'_i$ can reflect  the semantic  matching of words  and  images'  sub-regions, $Q'_i$  is  still  defined in the text embedding space and lacks necessary spatial structure information.  Thus, we design a convolution module ``ResBlock'' to  convert the word-context matrix $Q'_{i}$ to the  refined feature  $Q_i$. 
The spatial attention mask  $Mask_i^Q$  of the feature $Q_i$  is  defined as 

\begin{small}
	\begin{equation}\label{Spatial_Self_Attention_Mask1}
		Mask_i^Q= Sig.(Conv_{1 \times 1}^2 (ReLU(Conv_{3 \times 3}^1(Q_{i})))),
	\end{equation}
\end{small} 
We use $Q^+_{i} = Mask_i^Q \odot Q_{i}$ to express the  key spatial information of $Q_{i}$, and use  $Q^-_{i}=Q_{i}-Q^+_{i}$  to express the non-key information.

\subsubsection{Semantic Disentangling Loss}~\label{DDM_Section}
To drive  the SDM  to better distinguish between   the  key information and non-key information of $Q_i$ and $H_{i-1}$.
We further design a new Semantic Disentangling Loss (SDL) term. 
In   the  generation task,  the generated image distribution and the real image distribution are assumed to be the same type of distribution~\cite{Martin2017}. 
If the mean and variance of two distributions are the same, then two distributions are identical. 
Therefore, we use   the constraints of mean and variance to separate  key information  from non-key information of $Q_i$ and $H_{i-1}$, for  constructing the SDL loss.
Specifically, we  push the mean and variance of key information to approximate that of real images in  the latent space, and vice versa. 
The SDL on $H_{i-1}$  is  defined  as 

\begin{small}
	\begin{equation}\label{DDL_H_i}  
		\begin{aligned}
			\mathcal{L}_{SDL}^{H_i}=&SP(||\mu(H^+_{i})-\mu(H^*_{i})||-||\mu(H^-_{i})-\mu(H^*_{i})||)\\
			&+SP(||\sigma(H^+_{i})-\sigma(H^*_{i})||-||\sigma(H^-_{i})-\sigma(H^*_{i})||).
		\end{aligned}
	\end{equation}
\end{small}

Similarly,  the SDL   on the feature map $Q_i$  is  denoted  as  

\begin{small}
	\begin{equation}\label{DDL_Q_i}
		\begin{aligned}
			\mathcal{L}_{SDL}^{Q_i}=&SP(||\mu(Q^+_{i})-\mu(H^*_{i})||-||\mu(Q^-_{i})-\mu(H^*_{i})||)\\
			&+SP(||\sigma(Q^+_{i})-\sigma(H^*_{i})||-||\sigma(Q^-_{i})-\sigma(H^*_{i})||).
		\end{aligned}
	\end{equation}
\end{small}
Here, $\mu(\cdot)$ and $\sigma(\cdot)$ compute the mean and variance of feature maps within a batch,  the  $SP(x)=ln(1+e^{x})$.

The $H^*_i$ is the feature map of the corresponding real image  $I^*_i$.   We introduce  the Real Image Reconstruction Module (RIRM) shown in Fig.~\ref{TNNLS-2}  to  acquire  the real  image features $H^*_i$. The RIRM  contains  an encoder  and a decoder. The encoder  takes  the real image $I^*_i$ as  the input  and  outputs  the real image feature $H^*_i$. The decoder takes  the  real  image feature  $H^*_i$   and reconstruct   the  real  image  by  the reconstruction loss function $||RIRM(I^*_i)-I^*_i||_1$. Note that   the    decoder  and the generation modules  $G^o$  form  the Siamese Network,   which  can provide  high-quality  real  image features for SDM. 
The loss functions in SDM are summarized as 

\begin{small}
	\begin{equation}\label{DDM_total_loss}
		\mathcal{L}_{SDL_i}= \lambda_1 \mathcal{L}_{SDL}^{H_i}+\lambda_2 \mathcal{L}_{SDL}^{Q_i}+ \lambda_3 ||RIRM(I^*_i)-I^*_i||_1
	\end{equation}
\end{small}
Here,  $||RIRM(I^*_i)-I^*_i||_1$  is  the reconstruction loss  in  RIRM.

Finally, purified key features $Q^+_i$ and  $H^+_i$ are concatenated 
and then fused and upsampled (by using 
a ResBlock and Up-sampling module) to get the next-stage feature map $H_i$.

\begin{figure}[h!tb]
	\centering
	\includegraphics[scale=0.33]{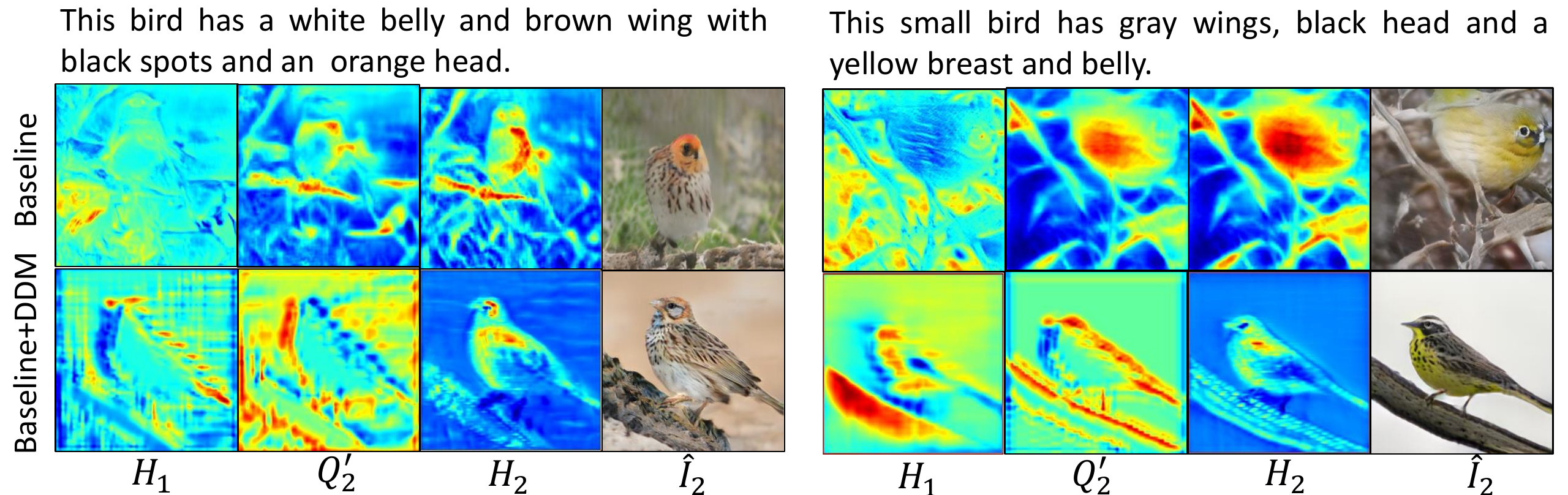}
	\caption{The image  feature  $H_1$ (Stage-1),  word-context matrix  $Q'_2$ (Stage-2), image  feature $H_2$ (Stage-2),   and  $H_2$'s  corresponding generated image $\hat{I}_2$ (Stage-2).  The examples in the first row  is from  our Baseline (Base.  AttnGAN~\cite{Xu2017AttnGAN}); The examples in the second row  is from  Baseline+SDM (Base.+SDM).   The warmer the color, the stronger the feature response. } 
	\label{TNNLS-3} 
\end{figure}

In  Fig.~\ref{TNNLS-3},  we  show   the  image  feature  $H_1$ (stage-1),  word-level context feature  $Q'_2$ (stage-2),  image  feature $H_2$ (stage-2), and the  synthesized image $\hat{I_2}$ (stage-2).   The first line is the visualization of Baseline (Base.) and the second line is the visualization of Baseline+SDM (Base.+SDM).

In the first line (Baseline) of Fig.~\ref{TNNLS-3}: Due to  abstract  and ambiguity of text descriptions, it is difficult for the generator  to accurately capture key semantics;  Some non-key semantics get mixed up in the generation process;  So, strange structures and details are easy to appear in the generated image features, such as $H_{1}$; the calculation of $Q'_2$ requires the participation of $H_{1}$, which also further  leads to $Q'_2$'s strange structure and details;  As a result, the structure of the generated image $\hat{I}_2$ is chaotic.
In the second line (Base.+SDM) of Fig.~\ref{TNNLS-3}:   Based on the selection strategy of key information driven by SDM, the  non-key  structural information on $H_{1}$ and $Q'_2$  can be better   filtered out.  Since this,  the structure  and  semantics of  image feature $H_2$ become more reasonable.  
	So,  the structure of the   synthesized  images $\hat{I}_2$  is also  reasonable.

\subsection{Distribution Normalization  Module (DNM)}~\label{DNM_Section}

The aforementioned SDM can help the \emph{generator} distill key semantic information  for  image  generation. 
But diverse visual information, messy background,  and   other  non-key  visual  information  in  images usually make  the image distribution more complicated. 
On the \emph{discriminator}'s side: such  complicated image distributions make the distinction of real and synthesized images harder; the discriminator may fail to effectively identify the synthesized image.

\begin{figure}[]
	\centering
	\includegraphics[scale=0.32]{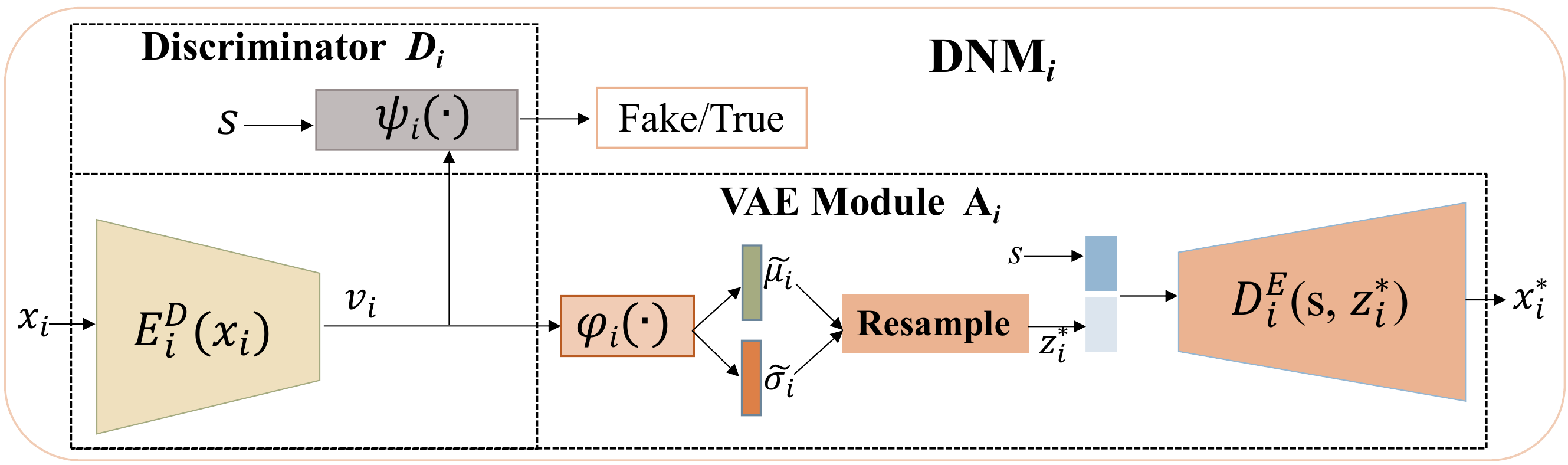}
	\caption{The architecture of the  DNM. The input $x$ is a generated image $\hat{I_i}$ or real image $I^*_i$. 
		The discriminator $D_i$ connects an encoder $E^D_i(\cdot)$ to a logical classifier $\psi_i(\cdot)$. 
		The VAE module consists of a variational encoder (that stacks $E^D_i(\cdot)$ and a variational sampling module $\varphi_i(\cdot)$), and a decoder ($D^E_i(\cdot)$).
	} 
	\label{TNNLS-4} 
\end{figure}

We know that the data normalization mechanism~\cite{Batch2015, ba2016layer, Dmitry16, Yuxin2018} can reduce the noise and internal covariate shift of data, and further  improve manifold learning efficiency in  the deep learning community.  
In the discriminant  stage of GAN, image sampling is firstly carried out from real image distribution and synthesized  image distribution. 
Secondly, the  discriminator determines whether the current input image is sampled from the real image distribution or  synthesized  image distribution, i.e. True or False.
Due to the diversity of images and the randomness of synthesized  images, the distributions of synthesized  images and real images   are  complicated.
Such complex  image   distribution  makes it difficult for the discriminator to distinguish whether the current input image is sampled from the real image distribution or generated image distribution.   
And  it  is  difficult  for the generator  to   align the generated distribution with the real image  distribution. 
Therefore,  it  is  necessary to reduce the complexity of the distribution. 
Normalization is an effective strategy for denoising and reducing complexity.
So, we expect to introduce the normalization of latent distribution.

As a generative model, Variational Auto-Encoder (VAE)~\cite{kingma2014autoencoding} can effectively  denoise the latent  distribution and reduce  complexity of  the distribution. In VAE, it assumes the latent embedding vector of an image  follow a Gaussian distribution $N(\tilde{\mu}, \tilde{\sigma})$, and then  normalizes the $N(\tilde{\mu}, \tilde{\sigma})$  to a standard normal distribution $N(0, 1)$.
Based on  the advantage  of image reconstruction in VAE,  the  normalized embedding vector  can   preserve key semantic visual  information.  
Thus, we build a VAE module in DNM to normalize the image latent distributions to help the discriminator better  distinguish between  the ``True'' image  and the  ``Fake'' image.

The structure of the $i$-th DNM is shown in Fig.~\ref{TNNLS-4}. DNM contains two sub-modules: the discriminator $D_i$ and the VAE module $A_i$. To simplify the notation, we omit the subscript $i$. $x$ can be a generated image $\hat{I}$ or a real image $I^*$.  
The \emph{discriminator} is composed of an encoder  $E^D(\cdot)$ and an logical classifier  $\psi(\cdot)$.   
$E^D(\cdot)$ encodes image $x$ into an embedding vector $v$. 
The embedding $v$ combined with text embedding $s$ is fed to the logical classifier $\psi(\cdot)$, which identifies if $x$ is a real or generated image. 
As mentioned above, diverse visual information, messy background,  and   other  non-key  visual  information in images make the distribution of embedding vectors $v$ complicated, and make the identification of $x$ harder. 
Thus, we adopt a \emph{VAE module} to normalize and  denoise the latent distribution of embedding vectors $v$. 
In  addition  to reducing the  complexity  of the image latent  distribution, 
using a VAE can also push the encoded image feature vector $v$ to  record important  image semantics (through reconstruction).

Our VAE module $A$ adopts a standard design architecture of Variational Auto-Encoder (VAE)~\cite{kingma2014autoencoding}.
As shown in Fig.~\ref{TNNLS-4}, 
$A$ has a variational encoder (which consists of an encoder $E^D(\cdot)$ and a variational sampling module $\varphi(\cdot)$), and a decoder $D^E(\cdot)$.

The information flow of the VAE module $A$ is as follows. 

\textbf{(1)} Given an image $x$, $x$ is first fed to  the encoder $E^D(\cdot)$, and $E^D(\cdot)$ outputs  the image latent embedding $v$.

\textbf{(2)} $\varphi(\cdot)$ infers the mean and  variance of $v$ and builds a Gaussian distribution $N(\tilde{\mu}(\varphi(v)), \tilde{\sigma}(\varphi(v)))$. $N(\tilde{\mu}(\varphi(v)), \tilde{\sigma}(\varphi(v)))$ is further normalized to a Normal distribution by the $KL(N(\tilde{\mu}(\varphi(v)), \tilde{\sigma}(\varphi(v))))||N(0,1))$. 
To make this procedure differentiable, the  re-sampling trick~\cite{kingma2014autoencoding} is adopted to get $z^*=z \cdot \tilde{\sigma}(\varphi(v))) + \tilde{\mu}(\varphi(v))$, $z \sim N(0,1)$.

\textbf{(3)} 
$z^*$ and the text embedding $s$ are concatenated, and then fed to the decoder $D^E(\cdot)$ to reconstruct image $x^*$.
Note that here the decoder takes both $s$ and $z^*$ for reconstruction, because the image generation here is conditioned on the text description.

\subsubsection{Distribution Adversarial Loss}

Following the alternate optimization mechanism of  GANs, our VAE module is trained together with the   discriminator. 
Based on the lower variational bound of the VAE~\cite{kingma2014autoencoding}, the loss  function of the VAE module in DNM$_i$ can be defined as

\begin{footnotesize}
	\begin{equation}\label{VAE_LOSS} 
		\begin{aligned}
			\mathcal{L}_{D_i^D} = &||\hat{I_i}-D^E_i(\varphi_i(E^{D}(\hat{I_i})),s)||_1+||I_i^*-D^E_i(\varphi_i(E^{D}(I_i^*)),s)||_1\\
			&+KL(N(\tilde{\mu_i}(\varphi_i(E^{D}(\hat{I_i}))), \tilde{\sigma_i}(\varphi_i(E^{D}(\hat{I_i})))))||N(0,1))\\
			&+KL(N(\tilde{\mu_i}(\varphi_i(E^{D}(I_i^*))), \tilde{\sigma_i}(\varphi_i(E^{D}(I_i^*)))))||N(0,1)).	
		\end{aligned}
	\end{equation}
\end{footnotesize}

In each generator's training step, the generated  images have an unnormalized distribution; while the distributions of real images have been normalized in the discriminator's training step. 
Hence, it is difficult for the generator to produce distributions to approximate the normalized real image distribution. 
Therefore, (1) we want to normalize the generated image distribution in the VAE module during the generator's training step. 
In addition, (2) we want to align the normalized generated distribution with the normalized real image distribution. 
To achieve the above two goals, we define a distribution consistency loss, i.e.

\begin{footnotesize}
	\begin{equation}\label{DCL}
		\begin{aligned}
			\mathcal{L}_{G_i^D} = 	&KL(N(\tilde{\mu_i}(\varphi_i(E^{D}(\hat{I_i}))), \tilde{\sigma_i}(\varphi_i(E^{D}(\hat{I_i})))))||N(0,1))\\
			&+||I_i^*-D^E_i(\varphi_i(E^{D}(\hat{I_i})),s)||_1, \\
		\end{aligned}
	\end{equation}
\end{footnotesize}
where the first term is designed for our first goal, and the second term is designed  for our second goal.

We denote two loss functions $\mathcal{L}_{G_i^D}$ and $\mathcal{L}_{D_i^D}$ as the Distribution Adversarial Loss (DAL) terms. 
In the discriminator's training stage,  $\mathcal{L}_{D_i^D}$ helps the discriminator  better distinguish the synthesized image from  the real image, and  better learn the distribution decision boundary between the generated versus real image latent distributions. 
In the generator's training stage,  $\mathcal{L}_{G_i^D}$  can help  the generator learn  and capture  the real image  distribution in  the  normalized  latent  space. 

Our DNM module, combining VAE and DAL, 
can effectively  reduce  the  complexity  of  distribution constructed by  the  image  embedding $v$, and enrich the  high-level semantic information of the  image  embedding $v$. 
Such a normalized embedding $v$ helps the discriminator 
better distinguish between the ``Fake'' image and the  ``True'' image. 
Consequently, the generator can also better align the generated distribution with the real image  distribution.

\subsection{Objective Functions in DR-GAN}~\label{Objective_Functions}

Combining the above modules, at the $i$-th stage of the DR-GAN, the Generative loss $\mathcal{L}_{G_i}$  and  Discriminative  loss $\mathcal{L}_{D_i}$ are defined as

\begin{footnotesize}
	\begin{equation}\label{eq10}
		\mathcal{L}_{G_i} = \underbrace{-\frac{1}{2}\mathbb{E}_{\hat{I_i}\sim P_{G_i}}[logD_i(\hat{I_i})]}_{\text{unconditional loss}}-\underbrace{\frac{1}{2}\mathbb{E}_{\hat{I_i}\sim P_{G_i}}[logD_i(\hat{I_i},s)]}_{\text{conditional loss}},
	\end{equation}
\end{footnotesize}
where the unconditional loss  is trained to generate high-quality images towards the real image distribution to fool  the discriminator, and  the conditional loss is trained to generate images to better match text descriptions. The $D_i(\cdot) = \hat{\psi}(E^D_i(\cdot))$   is  the  unconditional discriminator,   $D_i(\cdot, \cdot) = \psi(E^D_i(\cdot), s)$   is  the  conditional discriminator.  The   $\hat{\psi}(\cdot)$  and $\psi(\cdot)$  are  the unconditional   logical classifier   and    the conditional   logical classifier  respectively.

The discriminator $D_{i}$ is trained to classify the input image into the ``Fake'' or ``True'' class by minimizing the cross-entropy loss

\begin{scriptsize}
	\begin{equation}\label{eq11}
		\begin{aligned}
			\mathcal{L}_{D_i} = &\underbrace{-\frac{1}{2}\mathbb{E}_{I^*_i\sim P_{data_i}}[logD_i(I^*_i)] - \frac{1}{2}\mathbb{E}_{\hat{I_i}\sim P_{G_i}}[log(1-D_i(\hat{I_i})]}_{\text{unconditional loss}}+ \\
			&\underbrace{-\frac{1}{2}\mathbb{E}_{I^*_i\sim P_{data_i}}[logD_i(I^*_i,s)] - \frac{1}{2}\mathbb{E}_{\hat{I_i}\sim P_{G_i}}[log(1-D_i(\hat{I_i},s)]}_{\text{conditional loss}},
		\end{aligned}
	\end{equation}
\end{scriptsize}
where $I^*_{i}$ is from the realistic image distribution $P_{data}$ at the $i$-th scale, and $\hat{I_i}$ is from  distribution $P_{G_{i}}$ of the generative images at the same scale.

To generate realistic images, the final objective functions  in the  generation training stage ($\mathcal{L}_{G}$) and discrimination   training  stage  ($\mathcal{L}_{D}$) are respectively defined as

\begin{equation}
		\mathcal{L}_{G}=\sum_{i=0}^{m-1}(\mathcal{L}_{G_i} +\lambda_4\mathcal{L}_{G_i^D} + \mathcal{L}_{SDL_i})+ \alpha \mathcal{L}_{DAMSM},
\end{equation}  

\begin{small}
	\begin{equation}
		\mathcal{L}_{D}= \sum_{i=0}^{m-1}(\mathcal{L}_{D_i}+\lambda_5 \mathcal{L}_{D_i^D}).		
	\end{equation}
\end{small}
The  loss  function $\mathcal{L}_{DAMSM}$~\cite{Xu2017AttnGAN}  is  designed  to measure the matching degree between images and text descriptions. The DAMSM loss makes generated images better conditioned on text descriptions. 
The DR-GAN has three-stage generators ($m=3$) like the most recent GAN-based T2I  methods~\cite{Minfeng2019,  Bowen2020, Tingtingnips2019, 2019Tingting, Wenb19cvpr, Xu2017AttnGAN}.

\section{Experimental  Results}
\label{Experimental_Results}

\begin{table*}[h!tb]
	\centering
	\caption{\textbf{IS} $\uparrow$,  \textbf{FID} $\downarrow$, \textbf{MS} $\uparrow$, \textbf{R-Precision} $\uparrow$,    and  \textbf{Human Perceptual score (H.P. score)} $\uparrow$  by some SOTA GAN-based T2I models and our DR-GAN on the CUB-Bird and MS-COCO  test sets.   ${\dagger}$ indicates the scores are computed from	images generated by the open-sourced models.  $*$ indicates the scores are reported  in  DMGAN~\cite{Minfeng2019}. $**$ indicates the scores are reported  in  AttnGAN+O.P.*\cite{Tobias2019}.  Other results were reported in the original paper.   The Bold is the best result.
	} 
	
	\setlength{\tabcolsep}{1.5mm}{
		\begin{tabular}{c|c c c c c|c c c c c}
			\hline \hline
			\multirow{2}{*}{Method} &
			\multicolumn{5}{|c|}{$\textbf{CUB-Bird}$} &\multicolumn{5}{|c}{$\textbf{MS-COCO}$}  \\
			\cline{2-11} 	
			& IS $\uparrow$ & FID $\downarrow$ & MS $\uparrow$ &  R-Precision $\uparrow$ &  H.P. score $\uparrow$ & IS $\uparrow$ & FID $\downarrow$ & MS  $\uparrow$ &  R-Precision $\uparrow$  &  H.P. score $\uparrow$\\	
			\hline
			StackGANv2\cite{Han2018StackGANn} & $3.93 \pm 0.06$ & $29.64^{\dagger}$ & $4.10^{\dagger}$  &  -  & -  & $8.30 \pm 0.10$  & $81.59^{**}$  & -&  - & - \\ 
			AttnGAN \cite{Xu2017AttnGAN}   &  $4.36 \pm  0.02$  &  $23.98^{*}$  & $4.30^{\dagger}$   &  $52.62\%^{\dagger}$  & $16.14\%^{\dagger}$ &  $25.89 \pm 0.19$  &  $35.49^{*}$  & $23.71^{\dagger}$ &  $61.34\%^{\dagger}$ & $20.97\%^{\dagger}$\\ 
			Obj-GAN \cite{Wenb19cvpr}  & -  & -  &  -  &  -   & -  &   $30.29\pm 0.33$ & \textbf{25.64}  & - &  - & - \\ 
			DM-GAN \cite{Minfeng2019}  &  $4.75 \pm 0.07$  &  $16.09$  & $4.62^{\dagger}$   &  $59.21\%^{\dagger}$ & $32.47\%^{\dagger}$   &  $30.49 \pm 0.57$ & $32.64$  & $29.94^{\dagger}$ &  $70.63\%^{\dagger}$  & $37.51\%^{\dagger}$ \\ 
			RiFeGAN \cite{Juncvpr2020}  & \textbf{5.23 $\pm$ 0.09}  & -  &  -  &  -  & -  & $31.70$  & -  & - &  - & - \\ 
			\textbf{DR-GAN (Our)}   &  $4.90 \pm 0.05$ & \textbf{14.96} &  \textbf{4.86}  &  \textbf{61.17\%}  & \textbf{51.39\%} & \textbf{34.59 $\pm$ 0.51} &  $27.80$ & \textbf{33.96} &  \textbf{77.27\%}  & \textbf{41.52\%}\\ 
			\hline 
			\hline	
			
	\end{tabular} }
	\label{SOTA} 
\end{table*}

In this  section, we perform extensive experiments to evaluate the proposed DR-GAN.  Firstly, we compare our DR-GAN with other SOTA GAN-based T2I methods~\cite{Han2018StackGANn,Xu2017AttnGAN, Wenb19cvpr, Minfeng2019, Juncvpr2020}.   Secondly, we discuss the effectiveness of each new module introduced in DR-GAN: Semantic  Disentangling Module (SDM) and Distribution Normalization  Module (DNM). All the experiments are performed with one GTX 2080 Ti using the PyTorch toolbox.

\subsection{Experiment Settings}

\textbf{Datasets.}   We conduct experiments on two widely-used datasets,  CUB-Bird~\cite{WahCUB_200_2011}  and MS-COCO~\cite{Lin2014Microsoft}  datasets. The CUB-Bird~\cite{WahCUB_200_2011}  dataset contains $11,788$ bird images, and each  bird image has $10$ sentences  to describe the fine-grained visual details.   The MS-COCO dataset~\cite{Lin2014Microsoft} contains $80k$ training images and $40k$ test images, and each image has $5$ sentences to describe  the  visual information of the scene.  We pre-process and split the images on the two datasets by following the same setting in~\cite{Reed2016Generative}, \cite{Han2017StackGAN}.

\textbf{Evaluation.}  We from  three  aspects to compare our DR-GAN with other GAN-based T2I approaches: Image Diversity, Distribution Consistency, and  Semantic  Consistency. Each model generated $30,000$ images conditioning on the text descriptions from the unseen test set for evaluation. The $\uparrow$ means that the higher the value, the better the performance of the model, and vice versa.  

(I) \textbf{Image Diversity.}  Following almost all  T2I approaches, we adopt the fine-tuned Inception models~\cite{Han2017StackGAN} to calculate  the  Inception Score (\textbf{IS} $\uparrow$), which   measures  images diversity.

(II) \textbf{Distribution Consistency.}  We  use  the  Fr\'{e}chet Inception Distance (\textbf{FID} $\downarrow$)~\cite{Heusel2018GANs} and  Mode Score (\textbf{MS} $\uparrow$)~\cite{TongICLR2016} to  evaluate  the  distribution consistency  between  generated  images  and real  images.  The image features  in \textbf{FID} and  \textbf{MS}  are  extracted  by   a pre-trained Inception-V3 network~\cite{Szegedy2016Rethinking}.   

(III) \textbf{Semantic Consistency.} We  use  the  \textbf{R-precision $\uparrow$} and \textbf{Human Perceptual score (H.P. score $\uparrow$) }  to  evaluate  the semantic consistency between  the  text description and the synthesized  image. 

\textbf{R-precision.} Following ~\cite{Xu2017AttnGAN}, we also use  R-precision  to evaluate semantic  consistency. Given a pre-trained image-to-text retrieval model, we use generated images to query their corresponding text descriptions.
First, given generated image $\hat{x}$ conditioned on sentence $s$ and $99$ random sampled sentences $\{s_i^{'}: 1\leq i \leq 99\}$, we rank these $100$ sentences by the pre-trained image-to-text retrieval model. If the ground truth sentence $s$ is ranked highest, we count this as a successful retrieval. For all the images in the test dataset, we perform this retrieval task once and finally count the percentage of success retrievals as the R-precision score. Higher R-precision means greater semantic consistency.   

\textbf{Human Perceptual score (H.P. score).} To  get   \textbf{H.P.} score,  we randomly select $2000$ text descriptions on  CUB-Bird test set and $2000$ text descriptions on  MS-COCO test set.
Given the same text description, $30$ volunteers (not including any author) are asked to rank the images generated by different methods. 
The average ratio ranked as the best by human users is calculated to evaluate the compared methods.

\subsection{Comparison with state-of-the-arts}~\label{SOTA_Compare}

\begin{figure*}[h!tb]
	\centering
	\includegraphics[scale=0.215]{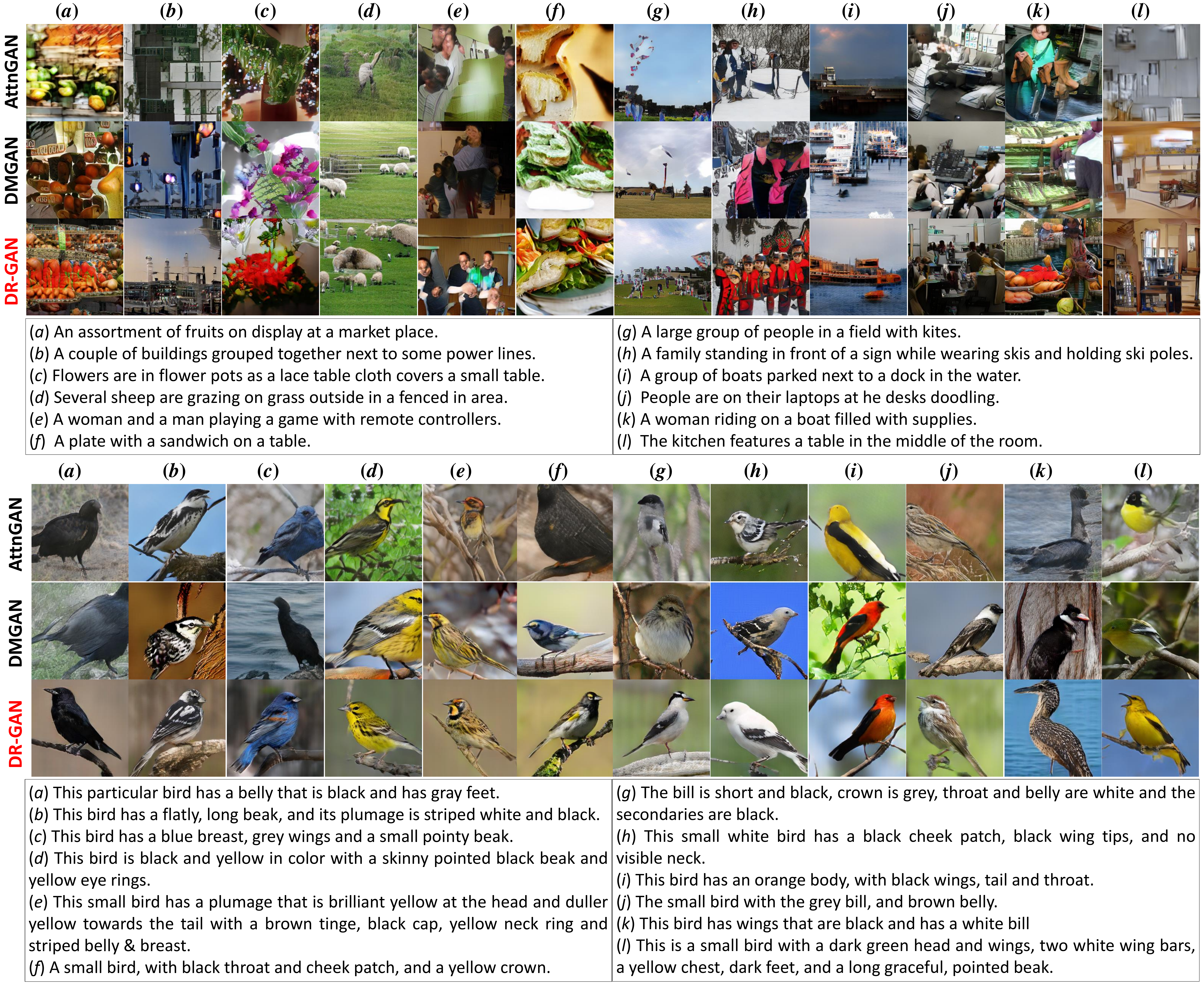}
	\caption{Images of $256\times256$ resolution are generated by our DR-GAN, DM-GAN~\cite{Minfeng2019}, and AttnGAN~\cite{Xu2017AttnGAN}  conditioned on text descriptions from the  MS-COCO (the upper part) and CUB-Bird (the bottom part) test sets. } 
	\label{TNNLS-5} 
\end{figure*}

\textbf{Image Diversity.}  We use \textbf{IS}  to evaluate the image diversity. 
As shown in Table~\ref{SOTA}, DR-GAN achieves the second-highest \textbf{IS} score on the CUB-Bird test set, and achieves the highest \textbf{IS} score  on the MS-COCO test set. 
The \textbf{IS} score ($5.23$) of RiFeGAN \cite{Juncvpr2020} is higher than that of ours ($4.90$) on the CUB-Bird dataset; but RiFeGAN \cite{Juncvpr2020} uses $10$ sentences to train the generator on the CUB-Bird dataset, while our DR-GAN only uses $1$ sentence following the standard T2I problem formulation. 
On the larger-scale and challenging MS-COCO dataset, RiFeGAN \cite{Juncvpr2020} uses $5$ sentences to train the model, but obtains a significantly lower \textbf{IS} score $31.70$ than that of our  DR-GAN ($34.59$). 
When the given database only contains images with a single sentence (which is common in practical tasks such as Story Visualization~\cite{StoryGAN2019},  Text-to-Video~\cite{Video2018}, and other  text-guided image  generation~\cite{yahuiACMMM2020}), RiFeGAN~\cite{Juncvpr2020}  can not be used. 
In contrast, methods such as our DR-GAN,  AttnGAN~\cite{Xu2017AttnGAN}, and  DMGAN~\cite{Minfeng2019} that only need one sentence per image can be used. 

\textbf{Distribution Consistency.}  We use  \textbf{FID}  and \textbf{MS} to evaluate the distribution consistency between the  generated image  distribution  and  the real image  distribution.  As  shown in Table~\ref{SOTA}, compared  with  these  GAN-based T2I methods,  our  DR-GAN achieves  the  competitive  performance   on the  CUB-Bird   and  MS-COCO test sets over  the \textbf{FID}  and the \textbf{MS}.  The  \textbf{FID} of   Obj-GAN \cite{Wenb19cvpr} is  lower than that  of our  DR-GAN.  This  is  because  Obj-GAN \cite{Wenb19cvpr} and its similar methods~\cite{Bowen2020, Tobias2019, Hong2018Inferring, Johnson2018Image, TobiasTPAMI2020}  require additional information, including the interesting object’s bounding
boxes and shapes, for training synthesizing.  This additional information  can  help  the generator  better capture  the object's layout. 
This additional information, although available in the MS-COCO dataset, is often
unavailable for other datasets such as the CUB-Bird dataset.
In general, producing more descriptions  or objects'  bounding
boxes and shapes  on a new database to train the generator is expensive. This limits its scalability and usability in more general text-guided image  generation. 
Thus,   our DR-GAN still  achieves  the  competitive  performance    in the distribution  consistency evaluation.

\textbf{Semantic Consistency.}  We  use  the  \textbf{R-precision} and  the  \textbf{Human Perceptual score (H.P.)}  to evaluate  the semantic  consistency.  As  shown  in Table~\ref{SOTA}, compared with AttnGAN and  DMGAN~\cite{Minfeng2019},  our DR-GAN  also  achieves  the best  performance on  the semantic  consistency evaluation on these  two  datasets.

\begin{table}[h!tb]
	\centering\small
	\caption{ The  performance  of applying SDM and DNM  on other  GAN-based T2I  methods  on CUB-Bird  test set.  The measures  include \textbf{IS}$\uparrow$, \textbf{FID} $\downarrow$   and  \textbf{MS} $\uparrow$.  AttnGAN$^{-}$  indicates  that  we  remove  the Word-level Attention Mechanism (WAM) in  AttnGAN~\cite{Xu2017AttnGAN}.  ${\dagger}$ indicates the scores are computed from	images generated by the open-sourced models.   Other results were reported in the original paper.}
	
	\setlength{\tabcolsep}{2mm}{
		\begin{tabular}{c|c c c  }
			\hline \hline
			\multirow{2}{*}{Method} &
			\multicolumn{3}{|c}{$\textbf{CUB-Bird}$}  \\
			\cline{2-4} 	
			& IS $\uparrow$  & FID $\downarrow$ & 	MS   $\uparrow$\\	
			\hline
			StackGANv2~\cite{Han2018StackGANn} &  $3.93 \pm 0.06$ & $29.64^{\dagger}$ & $4.10^{\dagger}$   \\ 
			StackGANv2+SDM+DNM    &  $4.58 \pm 0.04$ & $22.94$ &  $4.63$   \\ \hline 
			AttnGAN$^{-}$~\cite{Xu2017AttnGAN} &  $4.11 \pm 0.08$  &  $25.23^{\dagger}$  & $4.01^{\dagger}$  \\    
			AttnGAN$^{-}$+SDM+DNM   &  $4.65 \pm 0.06$  & $20.01$  & $4.54$  \\ \hline 
			DMGAN~\cite{Minfeng2019}&  $4.75 \pm 0.07$  &  $16.09$  & $4.62^{\dagger}$   \\   
			DMGAN+SDM+DNM    &  $4.84 \pm 0.04$  & $15.26$  &  $4.72$   \\ 
			
			\hline	\hline 
	\end{tabular} }
	\label{Generalization_SOTA} 
\end{table}

\begin{figure*}[h!tb]
	\centering
	\includegraphics[scale=0.212]{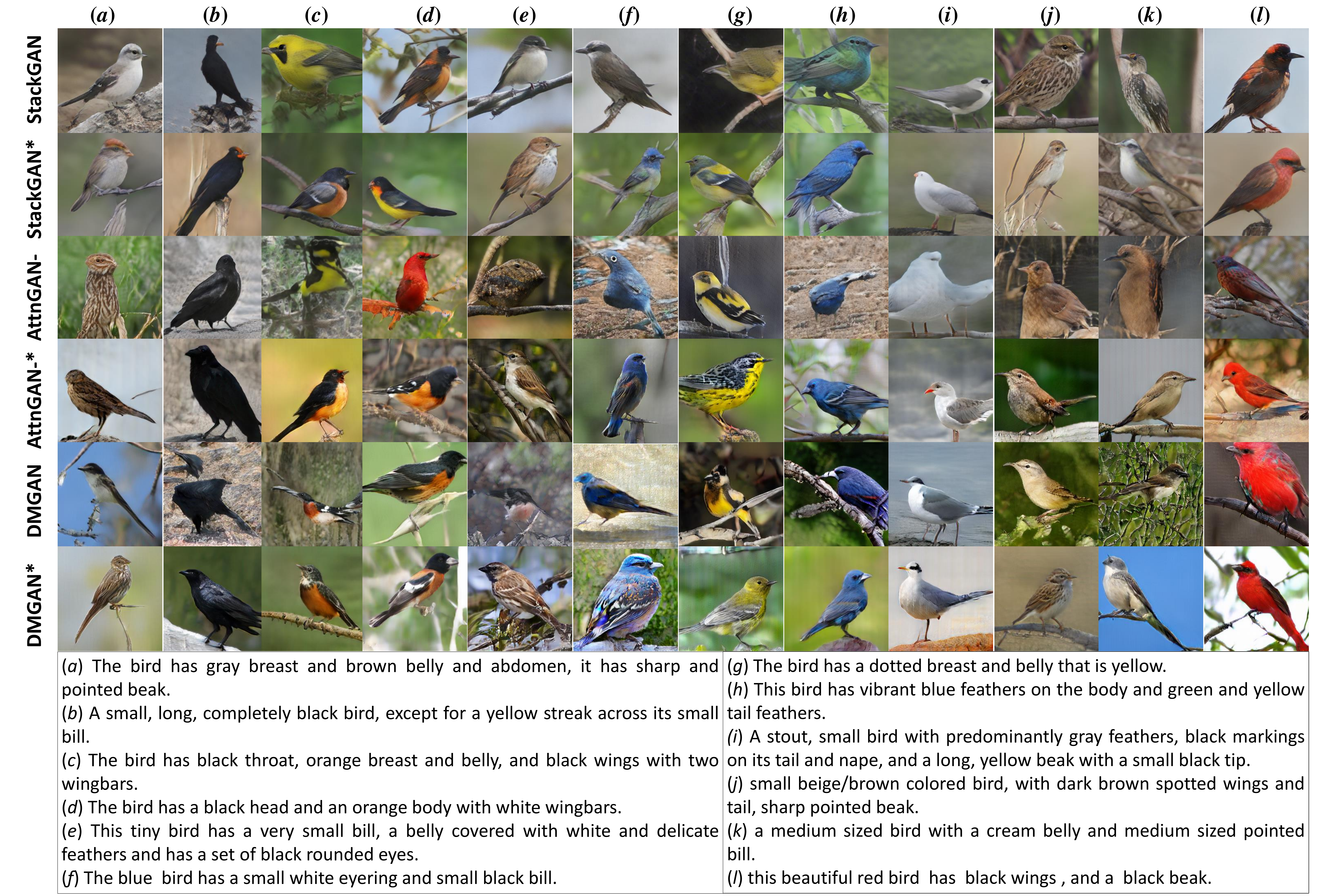}
	\caption{Images of $256\times256$ resolution are generated by  StackGAN v2~\cite{Han2018StackGANn},  StackGAN v2+SDM+DNM (StackGAN v2$^*$),  AttnGAN$^{-}$~\cite{Xu2017AttnGAN},  AttnGAN$^{-}$+SDM+DNM (AttnGAN$^{-*}$),  DMGAN~\cite{Minfeng2019}, and DMGAN+SDM+DNM (DMGAN$^*$)  conditioned on text descriptions from CUB-Bird  test set.} 
	\label{TNNLS-6} 
\end{figure*}
\textbf{Generalization.} To evaluate the generalizability of our proposed SDM and DNM, we integrate the SDM and DNM modules into several well-known/SOTA GAN-based T2I models including StackGANv2~\cite{Han2018StackGANn}, AttnGAN$^{-}$~\cite{Xu2017AttnGAN}, and  DMGAN~\cite{Minfeng2019}. Here, AttnGAN$^{-}$  denotes the AttnGAN~\cite{Xu2017AttnGAN} with its Word-level Attention Mechanism (WAM) removed. 
As shown in Table~\ref{Generalization_SOTA}, SDM and DNM can help all these GAN-Based T2I models achieve better performance on all three measures. 
It indicates that SDM and DNM can be used as general mechanisms in T2I generation task.

\textbf{Visualization.} We qualitatively  evaluate  our DR-GAN and some  GAN-based T2I  methods  by   image  visualization. As  shown in  Fig.~\ref{TNNLS-5}, compared with AttnGAN~\cite{Xu2017AttnGAN} and DMGAN~\cite{Minfeng2019},  our DG-GAN  can  synthesize  higher-quality  images  on CUB-Bird  and MS-COCO  test  sets. 

On the  CUB-Bird  test set:  (i) compared with our DR-GAN, the birds synthesized  by AttnGAN  and DMGAN contain  some  strange or incomplete details and structures; Because  SDM can better  distill  key information for image generation, our DR-GAN can perform better on the  structure generation; (ii) Due to the introduction of DNM, DR-GAN can better capture the  real images distribution. Therefore, compared with AttnGAN and DMGAN, the bird images generated by DR-GAN are more realistic and contain more full semantic details. In particular, the birds wings in Fig.-b-c-d-k have very detailed textures and full colors.
On the  MS-COCO test set:  the semantics of text description are rather sparse; it is very  difficult for the generator  to capture sufficient semantics to generate high-quality images; therefore, most current methods cannot synthesize  vivid  images;  compared with AttnGAN and DMGAN, our proposed DR-GAN is more reasonable in object  layout and  richer  in semantic details.

Besides,  in Fig.~\ref{TNNLS-6}, we have shown the synthesized  images of  StackGAN v2~\cite{Han2018StackGANn}, AttnGAN$^{-}$~\cite{Xu2017AttnGAN}, and DMGAN~\cite{Minfeng2019}, and  synthesized  images of  StackGAN v2+SDM+DNM (StackGAN v2$^*$),   AttnGAN$^{-}$+SDM+DNM (AttnGAN$^{-*}$), and DMGAN+SDM+DNM (DMGAN$^*$). 
As  shown in Fig.~\ref{TNNLS-6},    our  SDM and DNM   can  further  help  these  methods  (StackGAN v2~\cite{Han2018StackGANn}, AttnGAN$^{-}$~\cite{Xu2017AttnGAN}, and DMGAN~\cite{Minfeng2019})  improve the structure   and enrich the semantic details of synthesized  birds, so as to make the generated image more realistic.

\textbf{Model Cost.}
As shown in Tabel~\ref{model_size},  we  compare our  proposed  DR-GAN  with  other SOTA T2I  methods  under four model cost  measures including  Training Time, Training Epoch,  Model  Size,  and  Testing Time.  Using the MS-COCO dataset as an example, compared  with AttnGAN (Baseline) and DM-GAN,  our proposed DR-GAN is between AttnGAN and DM-GAN in terms of model cost while achieving the highest performance.

\begin{table}[h]
	\centering
	\caption{ The  Training Time, Training Epoch,  Model  Size, Testing Time    of our DR-GAN and other SOTA T2I methods   on the  MS-COCO dataset. }	
	\setlength{\tabcolsep}{0.1mm}{
		\begin{tabular}{c|c |c | c | c }
			\hline	\hline 
			$\textbf{Method}$     & Training Time  &    Training Epoch  & Model  Size  & Testing Time   \\   \hline
			AttnGAN (Base.)~\cite{Xu2017AttnGAN}    & $\sim 8 Days$  &  $120$ &  $\sim 55.5M$ &  $\sim 1200s$  \\  
			DM-GAN\cite{Minfeng2019}     & $\sim 14 Days$  &  $200$ &  $\sim 89.7M$ &  $\sim 1800s$  \\  
			DR-GAN(Ours)  &  $\sim 10 Days$ & $150$ & $\sim 73.2M$ & $\sim 1400s$   \\  \hline	\hline		
	\end{tabular} }
	\label{model_size}  	
\end{table}

\subsection{Ablation Strudy}

\subsubsection{Effectiveness of New Modules}

\begin{figure*}[h!tb]
	\centering
	\includegraphics[scale=0.215]{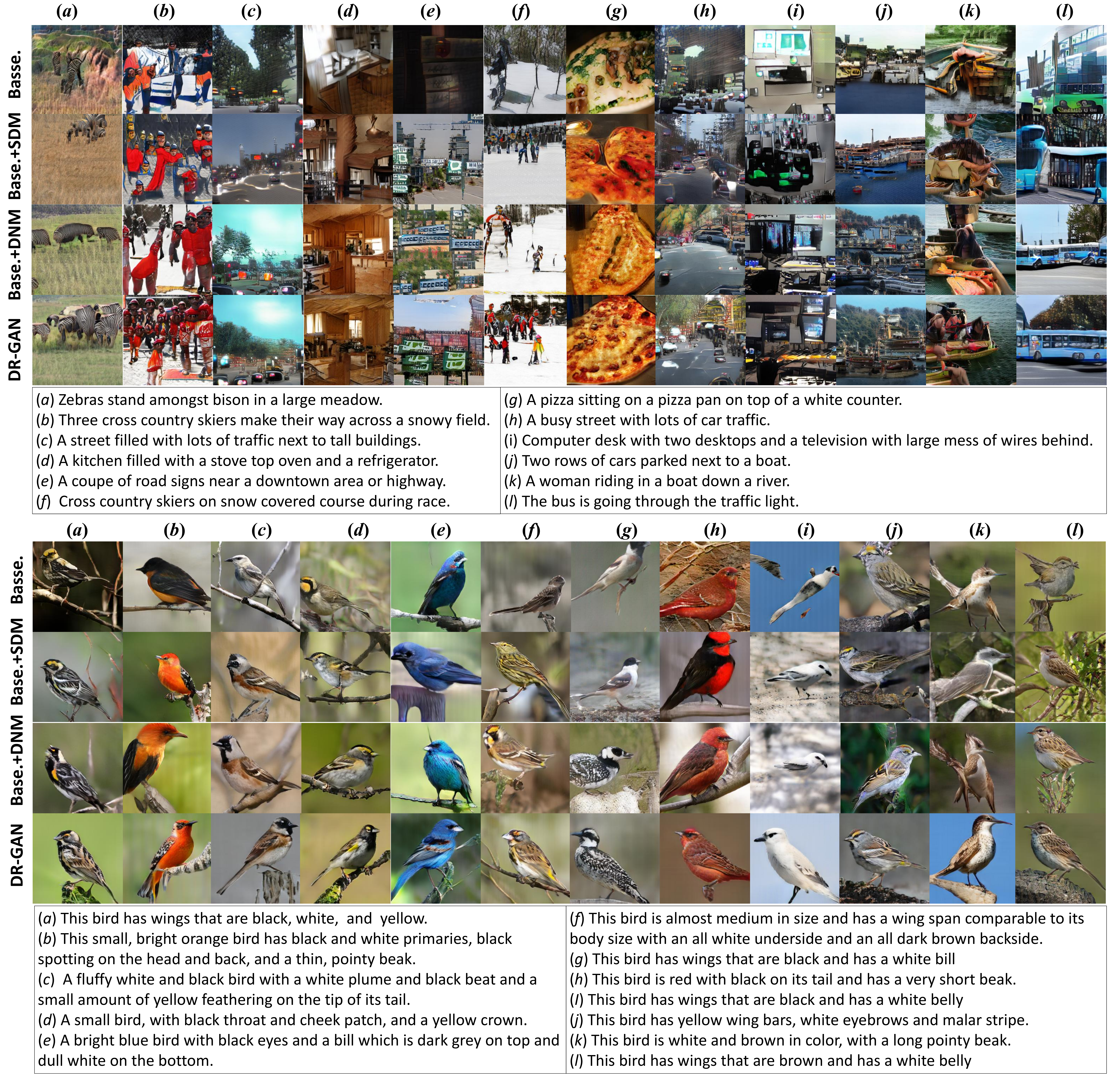}
	\caption{Images of $256\times256$ resolution are generated by our Baseline (Base.), Base.+SDM, Base.+DNM, and DR-GAN   conditioned on text descriptions from the CUB-Bird  test set.} 
	\label{TNNLS-7} 
\end{figure*}

\begin{figure*}[h!tb]
	\centering
	\includegraphics[scale=0.29]{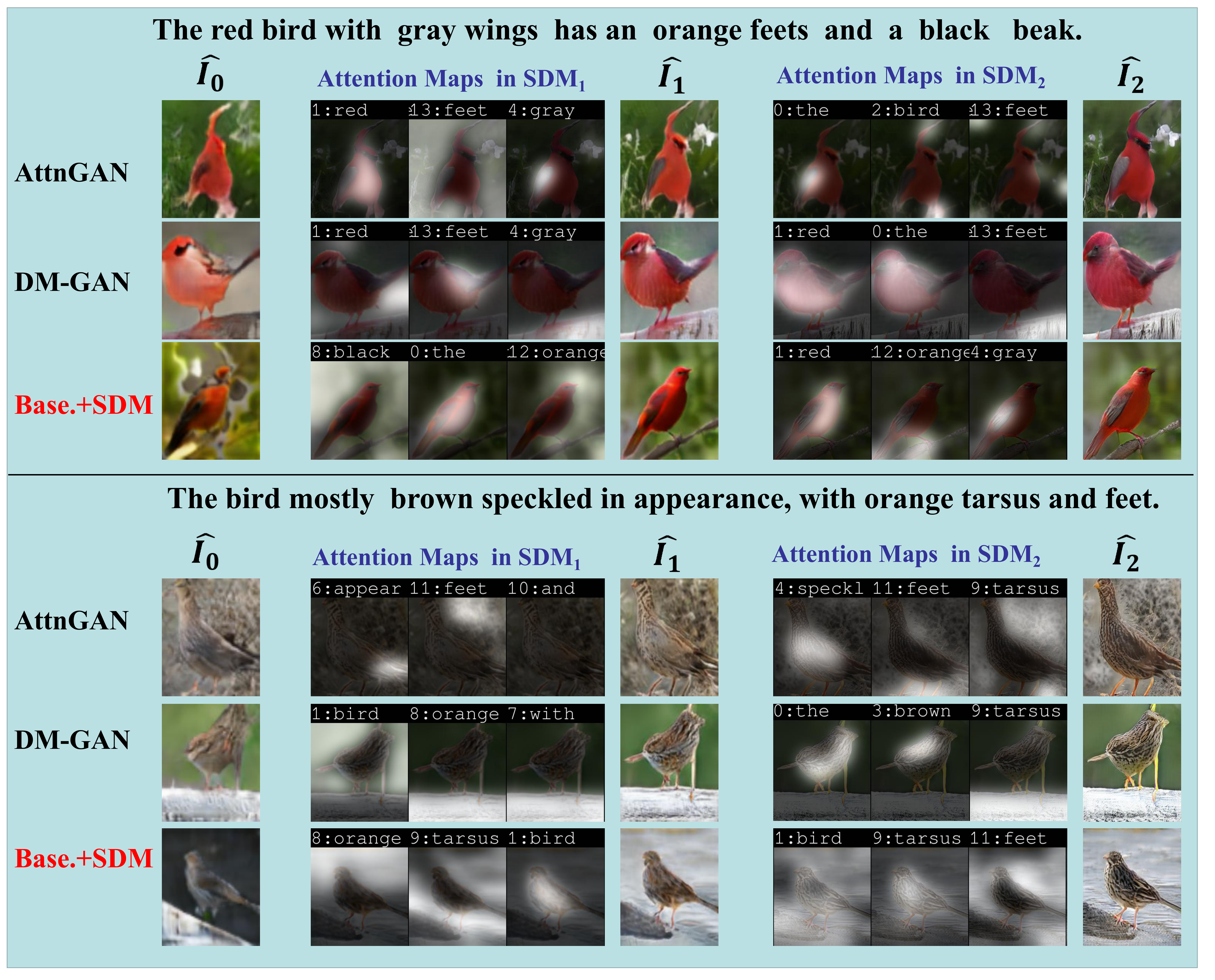}
	\caption{ The top-3 word guided attention maps and Synthesized  Images at different stages  from AttnGAN, DMGAN  and Base.+SDM (AttnGAN+SDM)  on   the  CUB-Bird test  dataset.} 
	\label{TNNLS-8} 
\end{figure*}

\begin{figure}[h!tb]
	\centering
	\includegraphics[scale=0.135]{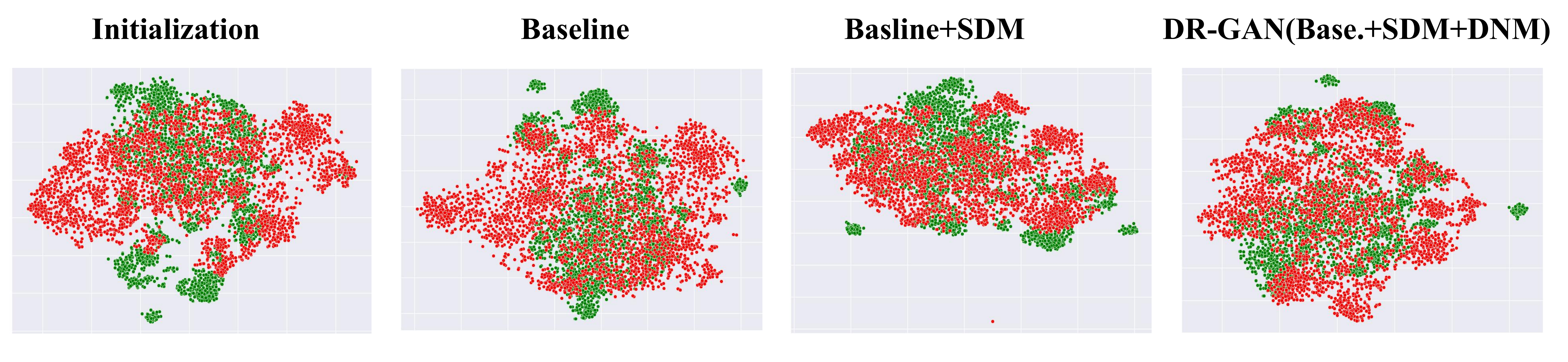}
	\caption{The visualization  results  of the synthesized images and real images under T-SNE~\cite{van2008}. The green scattered part is the real images, and the red scattered part is the generated images.} 
	\label{TNNLS-9} 
\end{figure}

\begin{table}[!t]
	\centering\small
	\caption{\textbf{IS}$\uparrow$, \textbf{FID} $\downarrow$   and  \textbf{MS} $\uparrow$  produced by combining different components of the DR-GAN  on the  CUB-Bird  and MS-COCO test sets. Our Baseline (Base.) is AttnGAN  \cite{Xu2017AttnGAN}.  DR-GAN=Base.+SDM+DNM.}
	
	\setlength{\tabcolsep}{0.4mm}{
		\begin{tabular}{c|c c  c|c c c }
			\hline \hline
			\multirow{2}{*}{Method} &
			\multicolumn{3}{|c|}{$\textbf{CUB-Bird}$} &\multicolumn{3}{|c}{$\textbf{MS-COCO}$}  \\
			\cline{2-7} 	
			& IS $\uparrow$ & FID $\downarrow$ & MS $\uparrow$ & IS $\uparrow$ & FID $\downarrow$ & MS $\uparrow$  \\	
			\hline
			Base. \cite{Xu2017AttnGAN}   &  $4.36 \pm  0.02$  &  $23.98$  & $4.30$     &  $25.89 \pm 0.19$  &  $35.49$  & $23.71$  \\
			Base.+SDM     &  $4.70 \pm 0.02$   & $15.50$  &  $4.62$   & $30.89\pm 0.57$   & $31.42$  &  $30.64$  \\
			Base.+DNM   & $4.79 \pm 0.03$  & $15.12$  &  $4.80$   & $31.26\pm 0.42$   & $29.73$ & $31.28$  \\ 
			DR-GAN    &  $4.90 \pm 0.05$ & $14.96$ &  $4.86$     & $34.59\pm 0.51$  & $27.80$ & $33.96$     \\ \hline 
			\hline	
	\end{tabular} }
	\label{ablation_table} 
\end{table}
In this  subsection, we evaluate the effectiveness of each new component qualitatively and quantitatively. The numerical results are documented in Table~\ref{ablation_table}. The visualization results  are  shown in Fig.~\ref{TNNLS-7}, Fig.~\ref{TNNLS-8}  and Fig.~\ref{TNNLS-9}.

\textbf{Quantitative results.} We evaluated the effectiveness of two new components, SDM and DNM, in terms  of three  measures. Our Baseline (Base.) is AttnGAN~\cite{Xu2017AttnGAN}.    As  shown in Table~\ref{ablation_table},  both  SDM  and DNM  can  effectively   improve  the performance  of   the baseline on these  two  datasets  over  three  measures. 

Firstly, we introduce  the SDM  into  the baseline (Base.), i.e. Base.+SDM.  
 As shown in Table~\ref{ablation_table}, Base.+SDM  leads to $7.80\%$ and $19.31\%$ improvement of \textbf{IS}, $35.36\%$ and $11.47\%$ improvement of \textbf{FID},  and $7.44\%$ and $29.23\%$ improvement of \textbf{MS}, on  the CUB-Bird and MS-COCO test sets respectively.  
 
Secondly, we introduce  the  DNM  into  the baseline (Base.), i.e. Base.+DNM.  
 As shown in Table~\ref{ablation_table}, Base.+DNM  leads to $9.86\%$ and $20.74\%$ improvement of \textbf{IS}, $36.95\%$ and $16.23\%$ improvement of \textbf{FID},  and $11.63\%$ and $31.92\%$ improvement of \textbf{MS}, on  the CUB-Bird and MS-COCO test sets respectively.  
 
Finally, when we  introduce the  SDM and the  DNM  into the baseline (Base.), i.e.  DR-GAN, our DR-GAN obtains $12.38\%$ and $33.60\%$ improvement over the baseline in \textbf{IS}, $37.61\%$ and $21.67\%$ improvement over the baseline in \textbf{FID},  and $13.02\%$ and $43.23\%$ improvement over the baseline in \textbf{MS},  on the CUB-Bird and MS-COCO test  sets  respectively. 
Our  DR-GAN  achieves $4.90$  and $34.59$  in the  term  of \textbf{IS}, and  achieves $14.96$  and $27.80$  in the  term  of \textbf{FID},   and achieves $4.86$  and $33.96$  in the  term  of \textbf{MS},  on the  CUB-Bird  and MS-COCO  test  sets respectively.

\textbf{Qualitative Results.}  We  also  qualitatively evaluate the  effectiveness  of each  component by image visualization (Fig.~\ref{TNNLS-7}, Fig.~\ref{TNNLS-8}, and  Fig.~\ref{TNNLS-9}).

In  Fig.~\ref{TNNLS-7},  we  can clearly  see that  the DNM  and SDM   can  effectively  improve  the quality of the generated images respectively.  
\textbf{SDM:} Compared with Baseline, the introduction of SDM can better improve bird body structure  on the CUB-Bird test set, and also improve target layout in complex scenes to some extent.  This is because SDM can directly extract key features and filter non-key features from the spatial perspective of feature maps.
Based on the constraints of real image feature distribution statistics,  SDM will try to filter out these non-key structural information which is not conducive to distribution learning. For COCO data, the excessive sparsity and abstraction of text semantics make the generator unable to generate vivid images. But with SDM, the overall layout and structure of generated images are significantly improved compared to the Baseline.

\textbf{DNM:} Compared with Baseline, the introduction of DNM can better improve visual representation and semantic expression of details to some extent.  This is because DNM normalizes the latent  distribution of real and generated images, which can drive the generator to better approximate the real image distribution. Therefore, generated images are better in terms of visual representation and detail semantics.
Compared with SDM, DNM lacks direct intervention on image features. Therefore, compared with Base.+SDM, the structure of images generated by Base.+DNM guidance is slightly worse.

\textbf{SDM+DNM:} When we introduce both SDM and DNM into the Baseline (Base.), i.e. DR-GAN. Based on the respective advantages of SDM and DNM, compared with Baseline, our proposed DR-GAN performs better in visual semantics, structure, and layout, and so on.

Besides, we present top-$3$ word guided attention maps  and $3$ stage generation images ( $\hat{I_0}$ $64 \times 6$, $\hat{I_1}$ $128 \times 128$, $\hat{I_2}$ $256 \times 256$)     from AttnGAN, DMGAN  and Base.+SDM (AttnGAN+SDM)  on   the  CUB-Bird test  dataset.  To facilitate display, we pulled the generated image pixels into images of the same size  in Fig.~\ref{TNNLS-8}.  For the  AttnGAN and DMGAN, we can observe that when the quality of the generated image $\hat{I_0}$ in the initial stage is very poor,  the  $\hat{I_0}$ affects the confusion in the word attention area.  Due to the lack of direct intervention of features, confused attention and image features continue to confuse $\hat{I_1}$ and attention maps in  the $SDM_2$. In contrast, the introduction of the proposed SDM  can gradually capture key information and filter  out  the non-key information for  image generation  in the subsequent stage. To this end, the attention information and the resulting images are gradually becoming more reasonable.

Finally,  we  use T-SNE~\cite{van2008}  to visualize the two-dimensional  distribution of 	 generated images and real images. 
As shown in  Fig.~\ref{TNNLS-9},  in the initial stage ($200$  epochs on Cub-Bird dataset),  the difference of the distribution between the generated image of Baseline and the real image is very  large. 
And the difference of the distribution between the generated image of Baseline and the real image is also large. 
When we introduce SDM into the Baseline, i.e. Baseline+SDM,  we observe a narrowing of the difference in the  distribution between the generated image and the real image. 
When we introduce DNM into the Baseline+SDM, i.e. DR-GAN, we observe a further narrowing of the difference in the two-dimensional distribution between the generated image and the real image. And DNM makes the scatter plot area more compact. 
Through the results of T-SNE visualization, we can see that DNM and SDM are effective for distribution learning.

\subsubsection{The validity of different parts of SDM.}
The \textbf{IS}$\uparrow$, \textbf{FID} $\downarrow$   and  \textbf{MS} $\uparrow$   of Baseline (Base.) are   $4.36 \pm  0.02$,   $23.98$  and   $4.30$ respectively.  
As shown  in  Table~\ref{SDM_Part},  we set $\lambda_3=0$ to mean that RIRM does not participate in SDM.
So, the SD loss  does not get  high-quality real image features.
The constraint of the whole distribution statistic will deviate from the feature distribution of the real image.
There will be a serious deterioration in image quality. 
Thus,  compared  with Base. and Base.+SDM,  the performance   of  Base.+SDM ($\lambda_3=0$)  is  bad.
When we remove the SDL, the performance of the model also deteriorates.
This is due to the lack of the necessary loss function to drive the attention mechanism to extract key information.
In all, SDM  can effectively improve the quality of model generation.

\begin{table}[!t]
	\centering 
	\caption{\textbf{IS}$\uparrow$, \textbf{FID} $\downarrow$   and  \textbf{MS} $\uparrow$  produced by  different parts in SDM on the  CUB-Bird  test set. }
	
	\setlength{\tabcolsep}{2mm}{
		\begin{tabular}{c|c c  c }
			\hline \hline
			\multirow{2}{*}{Method} &
			\multicolumn{3}{|c}{$\textbf{CUB-Bird}$}   \\
			\cline{2-4} 	
			& IS $\uparrow$ & FID $\downarrow$  & MS  $\uparrow$  \\	
			\hline   \hline
			Base. \cite{Xu2017AttnGAN}   &  $4.36 \pm  0.02$  &  $23.98$  & $4.30$    \\    
			Base.+SDM     &  $4.70 \pm 0.02$   & $15.50$  &  $4.62$   \\ 
			Base.+SDM w/o SDL  & $4.41 \pm 0.03$ & $23.56$ &  $4.38$   \\    
			Base.+SDM ($\lambda_3=0$) & $4.21 \pm 0.04$ & $27.34$ &  $4.11$   \\	 
			\hline 
			\hline	
			
	\end{tabular} }
	\label{SDM_Part} 
\end{table}

\subsubsection{Impact of self-encoding mode on performance. }
\begin{figure}[]
	\centering
	\includegraphics[scale=0.32]{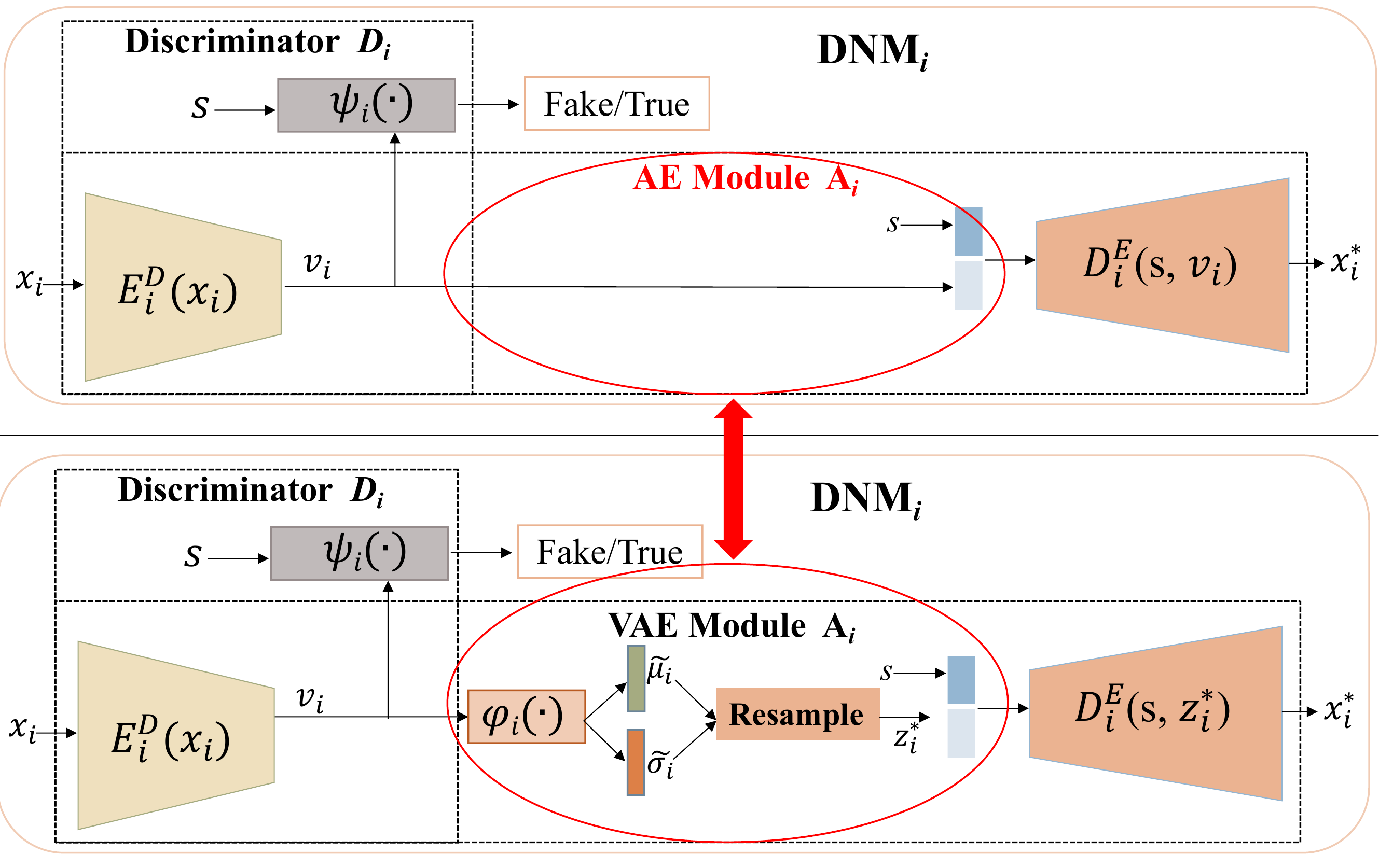}
	\caption{The architecture of the  DNM. The first case is VAE without variational sampling module, that is, general coding and decoding (AE).  The second case is the VAE framework adopted by us. 
	} 
	\label{TNNLS-10} 
\end{figure}

We modify the VAE module to observe their impact on model performance. The first case  in Fig.~\ref{TNNLS-10},  we  remove  the  variational sampling module $\varphi(\cdot)$) in VAE module.  Besides,  the  Distribution Adversarial Loss  is  rewritten as  Eq.~\ref{AE_LOSS} and Eq.~\ref{Ae_loss1}. 

\begin{footnotesize}
	\begin{equation}\label{AE_LOSS}
		\mathcal{L}_{D_i^D} = ||\hat{I_i}-D^E_i(\varphi_i(E^{D}(\hat{I_i})),s)||_1+||I_i^*-D^E_i(\varphi_i(E^{D}(I_i^*)),s)||_1
	\end{equation}
\end{footnotesize}

\begin{footnotesize}
	\begin{equation}\label{Ae_loss1}
		\mathcal{L}_{G_i^D} = ||I_i^*-D^E_i(\varphi_i(E^{D}(\hat{I_i})),s)||_1, 
	\end{equation}
\end{footnotesize}
Here we named such a model DNM$^*$. 
As  shown in Table~\ref{VAE_discussion},  we compare   Base.+DNM  and Base.+DNM$^*$  on  CUB-Bird dataset under  three  measures. 
Compared  with  Base.+DNM (contains VAE),  the performance of Base.+DNM$^*$ has significantly decreased.
However,  the  performance  of Base.+DNM$^*$ is  better than  that of Baseline.
This is because the image encoding and decoding in $AE$ can enhance the expression of image semantics in $v_i$.
In all, compared with the  general AE, adopting VAE can make the generator  obtain higher performance.
\begin{table}[!t]
	\centering 
	\caption{\textbf{IS}$\uparrow$, \textbf{FID} $\downarrow$   and  \textbf{MS} $\uparrow$  produced by Base.+DNM  and Base.+DNM$^*$  on the  CUB-Bird  test set. }
	
	\setlength{\tabcolsep}{2mm}{
		\begin{tabular}{c|c c  c }
			\hline \hline
			\multirow{2}{*}{Method} &
			\multicolumn{3}{|c}{$\textbf{CUB-Bird}$}   \\
			\cline{2-4} 	
			& IS $\uparrow$ & FID $\downarrow$  & MS  $\uparrow$  \\	
			\hline   \hline
			Base. \cite{Xu2017AttnGAN}   &  $4.36 \pm  0.02$  &  $23.98$  & $4.30$    \\    
			Base.+DNM$^*$     &  $4.53 \pm 0.05$   & $19.86$  &  $4.55$   \\ 
			Base.+DNM   & $4.79 \pm 0.03$  & $15.12$  &  $4.80$  \\  
			\hline 
			\hline	
			
	\end{tabular} }
	\label{VAE_discussion} 
\end{table}

\subsubsection{Discussion of semantic consistency strategies.}
We compare the performance of SDM with some outstanding semantic consistency strategies. 
Some  outstanding methods (such as  CSM-GAN$^*$~\cite{9359527},  SE-GAN$^*$\cite{thc2019}, MirrorGAN$^*$\cite{2019Tingting})  introduce  various  semantic constraints to improve the quality of image generation.
As  shown in Table~\ref{Semantics_IS},  we  present  the performance of the semantics  constraint  strategies in  these methods  under IS  score on the CUB-Bird dataset.
SD-GAN\cite{Guojun2019}  and SE-GAN\cite{thc2019} constrain semantic consistency between synthesized images and real images. 
CSM-GAN~\cite{9359527} constrains semantic consistency between the synthesized  images and  the text descriptions. 
MirrorGAN~\cite{2019Tingting} introduces semantic consistency between the text descriptions that the image is converted to and the real text descriptions. 
Compared with   these  methods,   distribution statistic constraints ( Baseline+SDM ) gain better performance  under IS score. 
We think that the randomness of images and texts leads to the instability of semantics constraints. 
The constraints of mean and variance can guide the learning direction of distributions for generators.
It helps the generator better align the generated image distribution with  the real image distribution.

\begin{table}[!t]
	\centering 
	\caption{Compare with other semantics consistent strategies  in some  outstanding T2I  methods  under  \textbf{IS} score. ``*'' represents the performance   the semantics  constraint  strategies in  these methods  under IS  score. }
	
	\setlength{\tabcolsep}{6mm}{
		\begin{tabular}{c|c}
			\hline \hline
			\multirow{2}{*}{Method} &
			\multicolumn{1}{|c}{$\textbf{CUB-Bird}$}   \\
			\cline{2-2} 	
			& IS $\uparrow$    \\	
			\hline   \hline
			Baseline~\cite{Xu2017AttnGAN}   &  $4.36 \pm  0.02$      \\    
			CSM-GAN$^*$~\cite{9359527}     &  $4.58 \pm 0.05$     \\ 
			SE-GAN$^*$~\cite{thc2019}  & $4.44 \pm 0.03$    \\    
			MirrorGAN$^*$~\cite{2019Tingting}  & $4.47 \pm 0.07$   \\
			SD-GAN$^*$~\cite{Guojun2019}  & $4.51 \pm 0.07$   \\	
			Baseline+SDM \textbf{(Our)}  &   $4.70 \pm 0.02$   \\ 
			\hline 
			\hline	
			
	\end{tabular} }
	\label{Semantics_IS} 
\end{table}

\subsection{Parametric Sensitivity Analysis.}
\begin{figure*}[h!tb]
	\centering
	\includegraphics[scale=0.58]{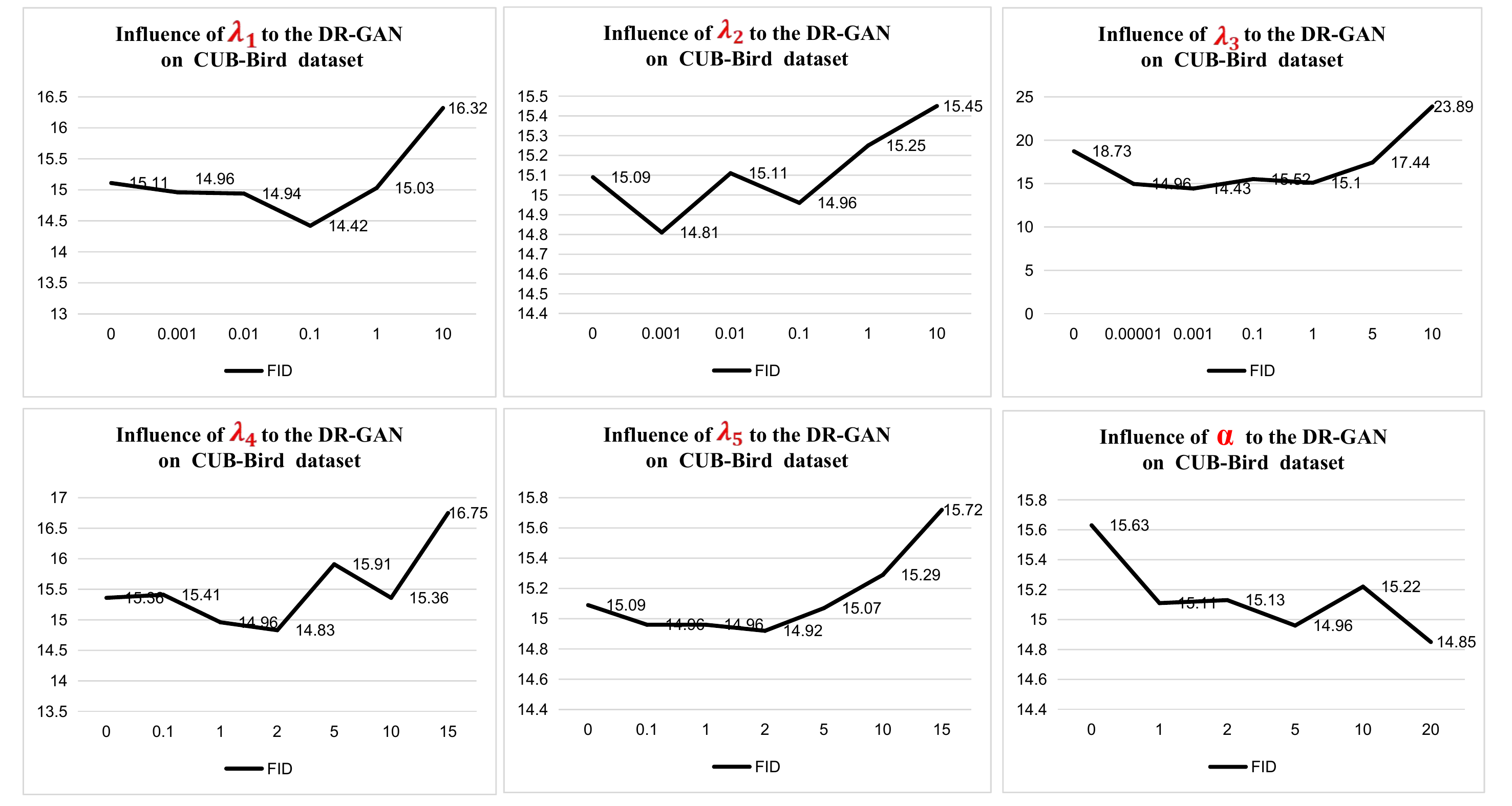}
	\caption{\textbf{FID} $\downarrow$   scores  of our  DR-GAN  under  different values of  the hyper-parameters  \textbf{$\lambda_1$, $\lambda_2$, $\lambda_3$, $\lambda_4$, $\lambda_5$, $\alpha$} on the CUB-Bird test  set. } 
	\label{TNNLS-11} 
\end{figure*}

In this  subsection, we   mainly  show  and analyze  the   sensitivity of the hyper-parameters  \textbf{$\lambda_1$, $\lambda_2$, $\lambda_3$, $\lambda_4$, $\lambda_5$, $\alpha$}.    In Table I to Table VII  (excluding Table III), the six parameters are  assigned to values when the DR-GAN performance is around the median value in Fig.~\ref{TNNLS-11}. So,  $\lambda_1=10^{-3}$, $\lambda_2= 10^{-1}$, $\lambda_3=10^{-5}$, $\lambda_4=1.0$, $\lambda_5=1.0$,  and $\alpha=5.0$.
The   FID  is an evaluation index to evaluate the quality of synthesized images on  the CUB-Bird dataset.  
The  FID  results of DR-GAN  under  different values of  the hyper-parameters  \textbf{$\lambda_1$, $\lambda_2$, $\lambda_3$, $\lambda_4$, $\lambda_5$, $\alpha$} are shown  in Fig.~\ref{TNNLS-11}.

\textbf{ The hyper-parameters $\lambda_1$, $\lambda_2$:} 
In the Semantic Disentangling Loss Eq. ~\ref{DDM_total_loss}, the loss  Eq.~\ref{DDL_H_i}   and  the  loss 	Eq.~\ref{DDL_Q_i}  are  proposed  to  drive  the SDM  better distill the   key information  from  image feature $H$ and word-level context feature $Q$  for image generation.  
The $\lambda_1$ and $\lambda_2$ are  important balance parameters in  the Semantic Disentangling Loss Eq. ~\ref{DDM_total_loss}.

As shown in Fig.~\ref{TNNLS-11}: (i) The Semantic Disentangling Loss has a great influence on the overall performance of DR-GAN.  When $\lambda_1=0$ or $\lambda_2=0$,  the FID score of DR-GAN increases, which means that the quality of the generated image deteriorates.  
(ii) As the values of $\lambda_1$ or $\lambda_2$ increase, the FID score also increases.    
The increase of weight means that model training pays more attention to the regression of distribution statistics.
The statistics reflect the overall information of the distribution.
The constraint of statistics is designed  to assist the generator to better approximate the real image distribution. 
The learning of accurate distribution requires GAN itself to approximate the real distribution based on implicit sampling. 
Therefore, the weight of SDL should be selected appropriately.
Based on the results shown in Fig.~\ref{TNNLS-11}, the  value range of $\lambda_1$  is $\{0.001, 0.01, 0.1 \}$, and  the  value range of $\lambda_2$  is $\{0.001, 0.1 \}$.

\textbf{The hyper-parameter   $\lambda_3$: } In the Semantic Disentangling Loss Eq. ~\ref{DDM_total_loss}, the  parameter $\lambda_3$   is designed  to adjust  the weight of  the  reconstruction loss   $ ||RIRM(I^*_i)-I^*_i||_1$. The  reconstruction loss   $ ||RIRM(I^*_i)-I^*_i||_1$  can provide  the real  image feature  $H^*$ for  other  terms in  Semantic Disentangling Loss.
When $\lambda_3=0$,  the reconstruction mechanism (i.e. RIRM) doesn't work.  So,  the performance of DR-GAN also drops significantly. 
With the removal of the loss, it is difficult for the SDM to match the valid mean and variance of real image features.
Since this, real image features  would  be mixed with more  semantic information irrelevant to image generation, or some important image semantic information would  be suppressed.
The  bad  real  image features  mislead the generator  to learn the bad image  distribution and make the quality of the generated image decline.
Besides, We found that performance decreased as the value  of $\lambda_3$ increased. 
Because   the  decoder in RIRM and the generation modules  $G^o$  are  the Siamese Networks. 
The purpose of this design is that the real image features $H^*$ and the generated image features $H$ can be mapped to the same semantic space.
The increase of weight will make the generation modules  $G^o$  pay more attention to the reconstruction of real images by real image features, and weaken the generation of generated images.
That is, the Siamese Networks are  prone to the  imbalance  between two feature sources in the training process.
In all, based on the results shown in Fig.~\ref{TNNLS-11}, the  value range of $\lambda_3$  is about $ [0.00001, 1]$.

\textbf{ The hyper-parameter $\lambda_4$, $\lambda_5$: }  The  hyper-parameter $\lambda_4$  balance the weight of  the  Eq.~\ref{DCL}   in  the generation stage loss  of DR-GAN;  The  hyper-parameter $\lambda_5$  balance the weight of    the   Eq.~\ref{VAE_LOSS} in  the generation stage loss  of DR-GAN; The  loss Eq.~\ref{DCL}  and loss form the  Distribution Adversarial Loss.   
When $\lambda_4=0$  or $\lambda_5=0$, the FID score increases, and the generated image quality decreases. 
This means that the two-loss terms of  Distribution Adversarial Loss can effectively improve the distribution quality of the generated image. 
When the value of the  $\lambda_4$  or $\lambda_5$ increases, the quality of the image distribution tends to decline. 
When the weight is too large, the image latent distribution will be over-normalized. At this point, the discriminant model becomes very powerful. 
Just like GAN's theory~\cite{Generative_2020}, a strong discriminator is not conducive to generator optimization. 
So, based on the results shown in Fig.~\ref{TNNLS-11}, the  value range of $\lambda_4$   and  $\lambda_5$  is  $[0.1 , 2]$.

\textbf{The hyper-parameter $\alpha$: } 
In the  training stage  of  DR-GAN, we  also  utilize the DAMSM loss~\cite{Xu2017AttnGAN} to make generated images better conditioned on text descriptions.  In Fig.~\ref{TNNLS-11},  we   show the performance  of  DR-GAN  based  on a different  value of the  hyper-parameter  $\alpha$. 
When $\alpha$, the FID score increases, and the generated image quality decreases.   This means that the constraints of semantic matching help to improve the quality of image generation. Besides,  when the value of $\alpha$ changes constantly, the performance of DR-GAN also changes moderately, and the overall performance is relatively good. Based on the results shown in Fig.~\ref{TNNLS-11}, the  value range of $\alpha$  is about $ [1 , 20 ]$.

\subsection{Limitation and Discussion}

\begin{figure}[h!tb]
	\centering
	\includegraphics[scale=0.155]{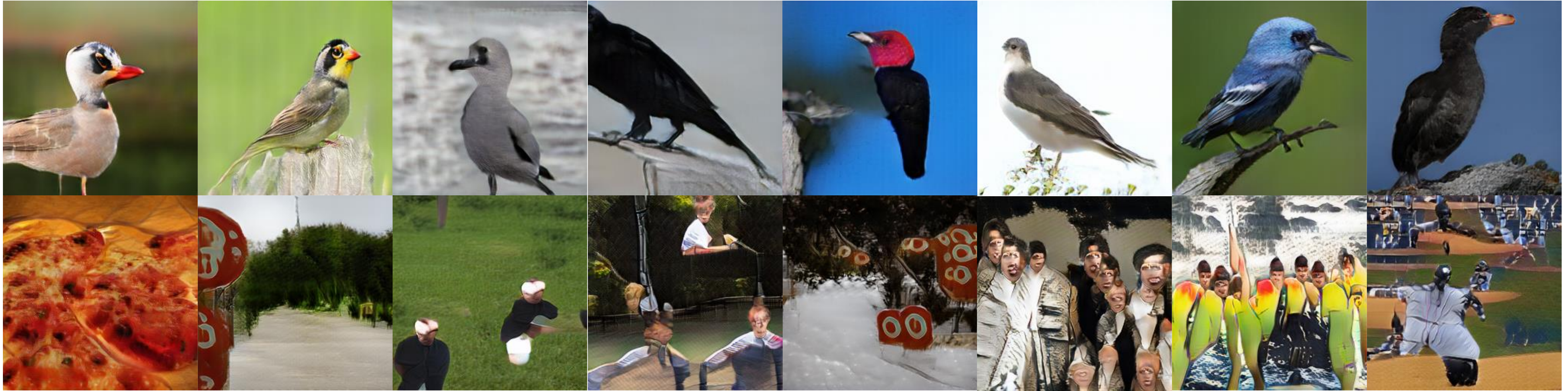}
	\caption{Failure cases generated by DR-GAN on  CUB-Bird test set (top row) and MS-COCO test set (bottom row).} 
	\label{TNNLS-12} 
\end{figure}

The experiments showed the effectiveness of our DR-GAN in T2I generation. 
However, there are a few failure cases.  
Some failure cases from the CUB-Bird data are shown in the top row of Fig.~\ref{TNNLS-12}.
Distribution normalization in SDM could sometimes lead to missing spatial local structures/parts such as heads, feet, and necks. 
Some failure cases from the MS-COCO  dataset are shown in the second row.   
Like in most other existing T2I models\cite{Xu2017AttnGAN, Wenb19cvpr, 2019Tingting, Bowen2020, thc2019, Guojun2019, Minfeng2019}, it is difficult for the text encoder to parse reasonable location information of objects in a scene if it is not specifically provided in sentences.
Then, the generator in DR-GAN tends to randomly place some objects that are sometimes unreasonable.  

We will explore the objects' location relationship from the text knowledge graph in our future work.
We will also explore the application of   SDM and DNM on other/broader GAN-based image generation tasks, such as Image-to-Image Translation~\cite{Jun-Yan2017iccv,  Phillip2017cvpr} and Virtual Try-On~\cite{Hancvpr2020, Assafcvpr2020}.

\section{Conclusions}
We proposed a novel Semantic Disentangling Module (SDM)  and  Distribution Normalization  Module (DNM) in the  GAN-based T2I model, and build a Distribution  Regularization Generative Adversarial Network (DR-GAN)  for Text-to-Image (T2I) generation. 
The SDM helps  the generator  better  distill  the  key information  and filter out  the  non-key  information  for image  generation. 
The DNM helps  GANs  better  normalize  and  reduce  the complexity  of  the image latent  distribution, and helps GAN-based T2I  methods  better  capture  the  real image distribution from  text feature distribution.
Extensive experimental results and analysis demonstrated the effectiveness of DR-GAN and  better  performance compared against previous  outstanding methods.  In  addition, the proposed  SDM and DNM  can further  help other   GAN-Based T2I models achieve better performance on the Text-to-Image  generation task. 
The  proposed  SDM and DNM  can be used as general mechanisms in Text-to-Image generation task.

\section*{Acknowledgments}

This work is supported by  National Key R\&D Program of China (2021ZD0111900),  National Natural Science Foundation of China (61976040), National Science Foundation of USA
(OIA-1946231, CBET-2115405), Chinese Postdoctoral Science Foundation (No. 2021M700303).
No conflict of interest: Hongchen Tan, Xiuping Liu, Baocai Yin, and Xin Li declare that they have no conflict of interest.

\ifCLASSOPTIONcaptionsoff
  \newpage
\fi

{\small
	\bibliographystyle{ieee_fullname}
	\bibliography{egbib}
}

\begin{IEEEbiography}[{\includegraphics[width=1in,height=1.25in,clip,keepaspectratio]{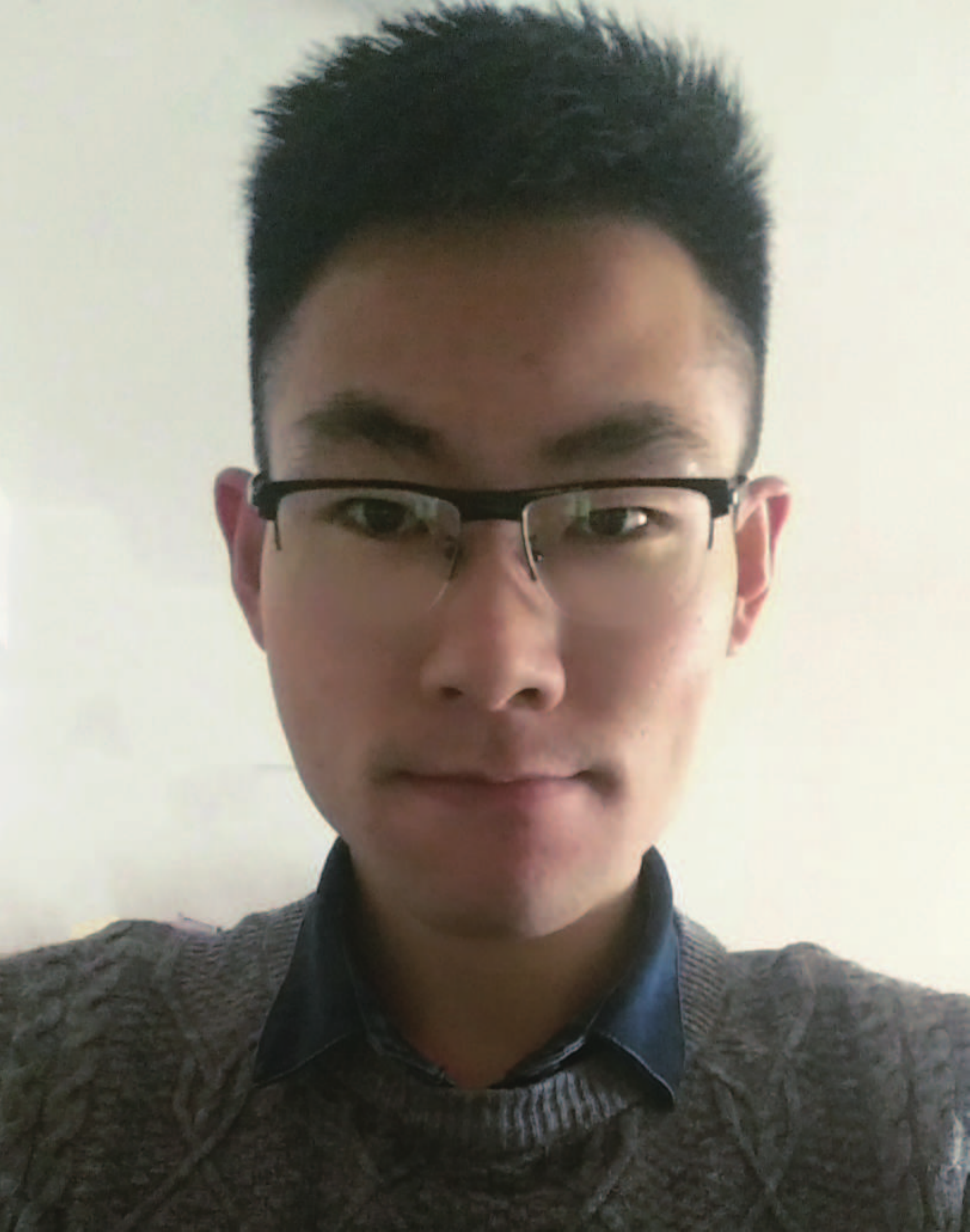}}]{Hongchen Tan} is a  Lecturer  of Artificial Intelligence Research Institute at Beijing University of Technology. He received Ph.D degrees in computational mathematics from  the Dalian University of Technology	in 2021. His research interests are  Person Re-identification,  Image Synthesis, and  Referring Segmentation.  Various parts of his work have been published in top conferences and journals, such as IEEE ICCV/TIP/TNNLS/TMM/TCSVT,  and Neurocomputing. 
\end{IEEEbiography}

\begin{IEEEbiography}[{\includegraphics[width=1in,height=1.25in,clip,keepaspectratio]{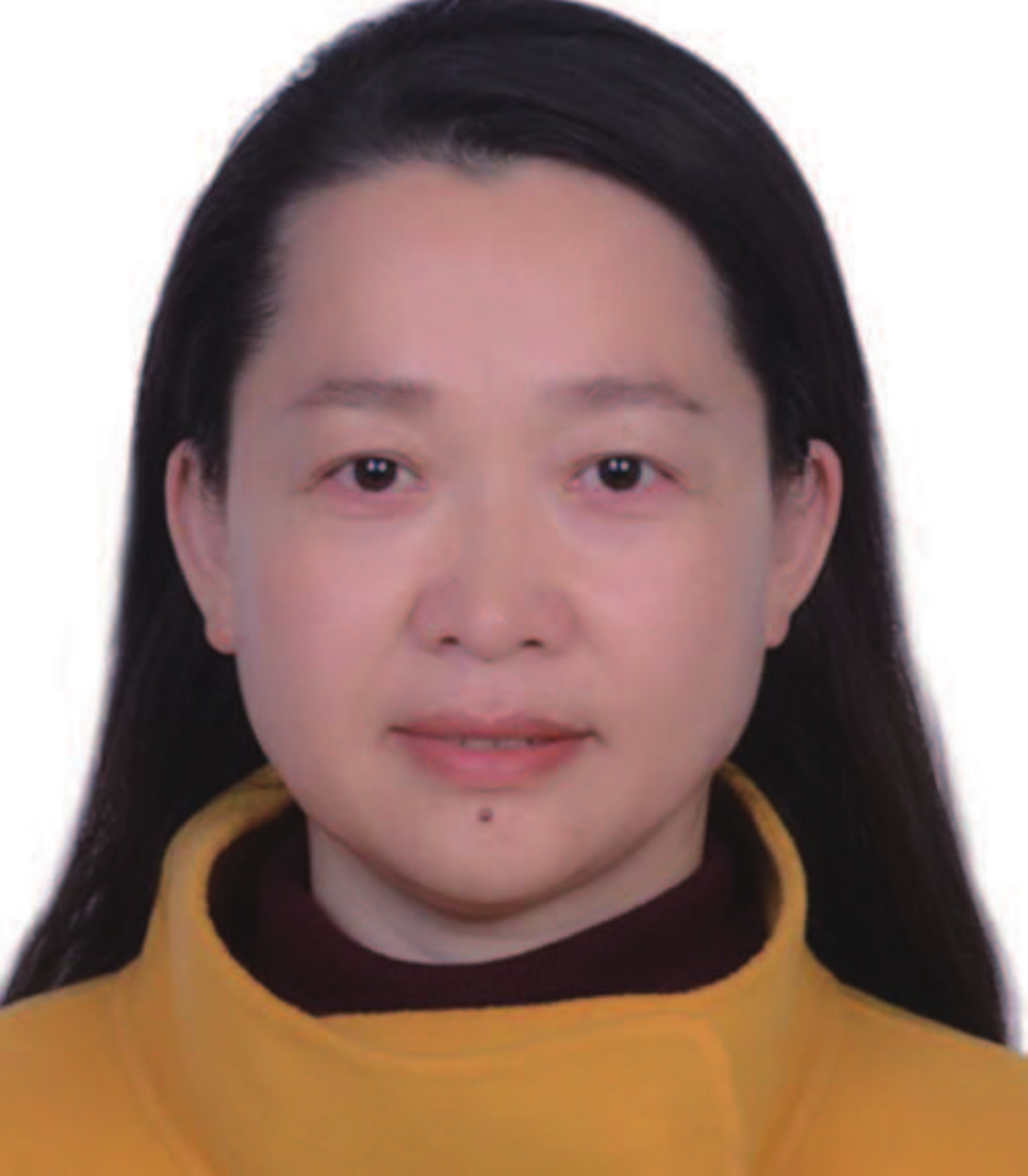}}]{Xiuping Liu}  is a Professor in School of Mathematical Sciences at Dalian University of Technology. She received Ph.D degrees in computational mathematics from Dalian University of Technology.   Her research interests include shape modeling and analyzing, and  computer vision.
\end{IEEEbiography}

\begin{IEEEbiography}[{\includegraphics[width=1in,height=1.25in,clip,keepaspectratio]{BaocaiYin}}]{Baocai Yin}  is  a	Professor  of Artificial Intelligence Research Institute  at  Beijing University of Technology. 	 He is	also a Researcher with the Beijing Key Laboratory of Multimedia and Intelligent Software Technology and the Beijing Advanced Innovation Center for Future Internet Technology. He received the M.S. and Ph.D. degrees in computational mathematics from the Dalian University of Technology, Dalian, China,	in 1988 and 1993, respectively.  His research interests include   multimedia, image processing, computer vision, and pattern recognition.
\end{IEEEbiography}

\begin{IEEEbiography}[{\includegraphics[width=1in,height=1.25in,clip,keepaspectratio]{XinLi}}]{Xin Li}  is a Professor at Division of Electrical \& Computer Engineering, Louisiana State University, USA. He got his B.E. degree in Computer Science from University of Science and Technology of China in 2003, and his M.S. and Ph.D. degrees in Computer Science from State University of New York at Stony Brook in 2005 and 2008.  His research interests are in Geometric and Visual Data Computing, Processing, and Understanding, Computer Vision, and Virtual Reality. 
\end{IEEEbiography}

\end{document}